\documentclass[11pt]{article}

\def\NAMEDVERSION{1} 
\ifdefined\NAMEDVERSION
\usepackage[preprint]{acl}
\else\ifdefined\FINALVERSION
\usepackage[final]{acl}
\else
\usepackage[review]{acl}
\fi
\fi

\usepackage{times}
\usepackage{latexsym}
\usepackage{textcomp}
\usepackage[T1]{fontenc}
\usepackage[utf8]{inputenc}
\usepackage{microtype}
\usepackage{inconsolata}
\usepackage{graphicx}
\usepackage{booktabs}
\usepackage{multirow}
\usepackage{amsmath}
\usepackage{tabularx}
\usepackage{placeins}
\usepackage{dblfloatfix}
\usepackage{tikz}
\usepackage{xcolor}
\usetikzlibrary{arrows.meta}

\graphicspath{{figures/}}

\title{Finding Errors Is Not Enough:\\
Selective Model Escalation for Entity-Targeted Sentiment Analysis}

\ifdefined\NAMEDVERSION
\author{Nishan Chatterjee\textsuperscript{1,2,3} \quad Matthew Purver\textsuperscript{1,5} \quad Antoine Doucet\textsuperscript{3,4} \quad Senja Pollak\textsuperscript{1} \\
\textsuperscript{1}Jo\v{z}ef Stefan Institute, Ljubljana, Slovenia \\
\textsuperscript{2}Jo\v{z}ef Stefan International Postgraduate School, Ljubljana, Slovenia \\
\textsuperscript{3}University of La Rochelle, La Rochelle, France \quad \textsuperscript{4}University of Ljubljana, Ljubljana, Slovenia \\
\textsuperscript{5}Queen Mary University of London, London, United Kingdom \\
\texttt{nishan.chatterjee@univ-lr.fr}}
\else
\author{Anonymous submission}
\fi

\begin{document}
\maketitle

\begin{abstract}
Improving a sentiment classifier with limited resources involves more than simply deciding how much data to collect or which model to train. A labelled collection may lack examples of an important class. Once trained, a classifier may benefit from another model even when that model is weaker on its own. We study these decisions in document-level Entity-Level Sentiment Analysis (ELSA), primarily using South Slavic news. We first compare acquisition orders as collections grow. We then use uncertainty to choose which predictions to revisit, how many to process, and whether an alternative encoder or local LLM can correct them. Finally, we select complete routing policies on development data for held-out evaluation. Minority-prioritised acquisition helps at some very small training sizes, but not consistently thereafter. On SentiCoref, an oracle correcting all errors in the least-confident 10\% without harming correct predictions would add 9.14 Macro-F1 points;  however the best evaluated action gains only 0.90. Encoder averaging helps on AspectBench and NewsMTSC but harms SentiCoref, where selective LLM assistance is more promising. Both development-selected policies route 30\% of records to the encoder-assisted LLM, yielding observed gains of 1.02 points on SentiCoref (interval includes zero) and 1.71 on NewsMTSC, at substantially different inference costs.
\end{abstract}

\section{Introduction}
News articles can express different attitudes towards several people and organisations. An article may praise a programme, criticise the responsible minister, and quote an opposition party without endorsing its position. A single article-level sentiment label cannot capture all these differences. Our task is therefore to predict the sentiment towards a supplied entity across the article as a whole. We study this as document-level Entity-Level Sentiment Analysis (ELSA) \citep{ronningstad-etal-2022-entity}, within the broader study of target-dependent and aspect-based sentiment analysis \citep{hamborg-donnay-2021-newsmtsc,pei2019targetedsentimentanalysisdatadriven,pontiki-etal-2014-semeval}. Because the entities are already annotated, target and aspect extraction fall outside our task.

Improving such a classifier involves deciding where additional effort is needed. South Slavic news collections are relatively small and unevenly distributed across sentiment classes, languages, and scripts. When a sentiment class is scarcely represented, obtaining more examples of that class may help more than simply enlarging the collection. Continued prioritisation can also leave other parts of the data underrepresented. The composition of a growing collection therefore matters alongside its size.

Once a model has been trained, improvement need not mean replacing it with a larger or stronger classifier. Long articles contain repeated entity mentions and opinions attributed to different speakers, so some predictions may deserve another look. Different models can also make different errors. A second model may therefore help on selected predictions even if its overall performance is weaker than that of the primary classifier. The challenge is to identify those predictions, decide how many to process, and choose an action that corrects more mistakes than it introduces. Additional processing also carries a computational cost. We study this through post-hoc selective model routing, related to learning to defer but without training the complete system end to end.

\citet{chatterjee2026evaluating} introduced AspectBench and provided the modelling foundation for long South Slavic news. Their study compared model architectures and learning curves, documented differences across languages and scripts, and found supervised encoders to be the strongest standalone classifiers among the evaluated systems. It also showed that confidence-selected assistance from a local LLM could help despite the evaluated LLMs' weak standalone performance.

These results leave four questions. Do composition and acquisition order matter as the collection grows? Which uncertainty signals identify errors, and at what computational cost? Does another model improve the same selected predictions without introducing too many new errors? Finally, can development data choose the signal, processing fraction, and action before test labels are inspected?

We organise these questions as four linked decisions: \emph{Acquire}, \emph{Detect}, \emph{Resolve}, and \emph{Select}. Acquire examines how an existing labelled collection grows. The other three decisions begin with a trained primary classifier and follow a sequence from identifying risky predictions, through comparing corrective actions, to choosing a complete policy before test evaluation. They belong to the same development process but are evaluated separately: we do not propagate every acquisition history through routing or jointly optimise annotation and inference budgets.

Beyond this experimental framing, our contributions are:
\begin{enumerate}
    \item \textbf{Acquisition-order evidence.} We extend the earlier learning-curve analysis by \cite{chatterjee2026evaluating} with retrospective comparisons of how class-prioritised, random, and chronological acquisition histories affect a growing collection. Although, this is not an online active-learning experiment.
    \item \textbf{A comparative corrective-policy space.} We evaluate uncertainty signals, selected fractions, and competing encoder and LLM actions, including matched-queue comparisons that distinguish error-detection headroom from corrections and harms actually realised by available models.
    \item \textbf{Development-to-test policy selection.} We select a complete signal--fraction--action policy using source-disjoint development predictions, freeze the choice before held-out test evaluation, and report its observed gain, empirical selection regret, and additional compute. This tests policy selection rather than retrospectively reporting the test-best action.
    \item \textbf{External and exploratory complementary-policy evidence.} We extend the corrective-action comparisons beyond AspectBench to SentiCoref and NewsMTSC, exposing dataset-dependent differences between encoder and LLM assistance. We also test action-specific uncertainty preferences and learned per-record corrective-action routers; the mixed results do not establish a consistently superior complementary routing rule.
\end{enumerate}

\section{Related Work}
Before training, active learning and imbalance-aware acquisition ask which unlabelled records are worth annotating \citep{ein-dor-etal-2020-active,fairstein-etal-2024-class,zhang-plank-2021-cartography-active}. Class-aware selection can help find rare examples, although repeatedly prioritising them may narrow the range of contexts represented in the collection. Annotation effort also extends beyond label counts to sampling, guideline development, supervision, and correction \citep{haertel-etal-2008-assessing,schepers-etal-2025-price}. Our experiment reconstructs alternative acquisition orders within an existing labelled pool. It therefore examines how the composition of a growing training set affects learning, without implementing an online strategy for finding new examples. Random ordering remains an important baseline because active-learning gains do not always transfer across settings \citep{lowell-etal-2019-practical}.

After training, selective classification asks whether predictions likely to be wrong can be identified while retaining coverage on the remainder \citep{geifman2017selective}. Monte Carlo dropout provides several ways to estimate uncertainty and rank risky cases \citep{pmlr-v48-gal16}. Calibration addresses a related but different question: whether reported confidence corresponds to empirical correctness \citep{pmlr-v70-guo17a}. A score can rank errors well without being calibrated.

Learning-to-defer systems and model cascades study when an input should pass to another decision-maker or model \citep{madras2018predict,varshney-baral-2022-model,narasimhan-etal-2022-posthoc,mao-etal-2023-two-stage}. A primary-model risk signal need not rank a second model's benefit: both may fail on the same case, or the second may damage a correct decision. Cost-aware cascades and routers address model use under constraints, while RouteLMT learns expected marginal gain \citep{chen-etal-2024-frugalgpt,luo-etal-2026-routelmt}. We compare actions on the same predictions, then test development selection of the complete rule, fraction, and action \citep{cawley-talbot-2010-overfitting}.

The direct empirical basis remains \citet{chatterjee2026evaluating}: AspectBench, model and ensemble comparisons, learning curves, and local-LLM deferral. We extend that evidence from retrospective comparisons to acquisition order, alternative uncertainty signals, matched corrective actions, and development-selected policies. The earlier deferral results did not systematically compare corrective actions on matched queues or select the complete routing policy before test evaluation.

\section{Research Questions and Study Design}
RQ1 concerns the collection before full-scale training; RQ2--RQ4 begin with a trained primary classifier. We evaluate these branches separately rather than propagating every acquisition history into routing.

\noindent\textbf{RQ1, Acquire.} How should a growing labelled collection be assembled, and when does targeted acquisition outperform broader sampling?

\noindent\textbf{RQ2, Detect.} Which uncertainty signals and selection rules identify predictions worth revisiting, and how much correction opportunity do they expose?

\noindent\textbf{RQ3, Resolve.} When the same predictions are selected, which corrective actions help, which introduce new errors, and how does their usefulness depend on the selected records?

\noindent\textbf{RQ4, Select.} Can development data select a useful complete policy, and does that policy remain useful on held-out records and under controlled population resampling?

Throughout RQ2--RQ4, we use \emph{E1} for the primary supervised encoder, whose prediction is the system's starting point, and \emph{E2} for a separately trained encoder that can provide an alternative prediction. We use \emph{LLM} for the local generative model that can instead reconsider an E1 prediction. All reported LLM comparisons use the instruction-tuned, 4-bit Gemma 3 27B QAT checkpoint, served locally through vLLM \citep{gemmateam2025gemma3technicalreport}. Our audit of the stored outputs found no missing queries or malformed responses in the comparisons reported here. Appendix Figure~\ref{fig:rq-structure} shows the independent RQ1 branch and the Detect--Resolve--Select sequence.

\subsection{Datasets and task conversion}

Table~\ref{tab:data} combines the two AspectBench collections with two external evaluations. AspectBench contains long South Slavic news: Slovenian is separate, while HBS comprises Serbian (SR), the dataset's Serbo-Croatian subset (SH), Bosnian (BS), and Croatian (HR) \citep{chatterjee2026evaluating,chatterjee2026aspectbenchhbs}. We map SentiCoref's five-level Slovenian sentiment to three classes and split by source document \citep{zitnik2019senticoref,zitnik2022target}; NewsMTSC supplies its official English political-news test \citep{hamborg-donnay-2021-newsmtsc}. These external datasets test recurrence beyond AspectBench but do not isolate language effects. Appendix~\ref{sec:detailed-protocol} gives conversion details.

\captionsetup[table]{skip=2pt}
\setlength{\textfloatsep}{3pt plus 1pt}
\begin{table}[t]
\centering
\small
\setlength{\tabcolsep}{4.5pt}
\begin{tabular}{@{}lrr@{}}
\toprule
Dataset & Test $n$ & \shortstack{Neg. / Neu. / Pos. (\%)} \\
\midrule
AspectBench HBS & 12,407 & 7.5 / 49.5 / 43.0 \\
AspectBench SL & 7,740 & 1.2 / 82.6 / 16.2 \\
SentiCoref SL & 2,155 & 12.9 / 75.1 / 12.0 \\
NewsMTSC EN & 1,146 & 37.4 / 39.7 / 22.9 \\
\bottomrule
\end{tabular}
\caption{Test article--entity counts and sentiment shares (\%). HBS groups Serbian, Serbo-Croatian, Bosnian, and Croatian; SL and EN are Slovenian and English.}
\label{tab:data}
\end{table}

AspectBench is already labelled: RQ1 reveals its article--entity records in different orders. HBS uses BERTi\'{c} as E1 and XLM-R as E2; AspectBench SL and SentiCoref use SloBERTa and XLM-R; NewsMTSC uses XLM-R and mDeBERTa-v3. These pairings follow the earlier benchmark and external evaluations \citep{chatterjee2026evaluating}. Macro-F1 is our principal metric; we also report accuracy, correction and harm counts, and quadratic weighted kappa (QWK), which penalises larger ordinal disagreements more heavily.

\subsection{Acquisition histories}
We reveal the fixed AspectBench training and validation pools in random order, minority-prioritised order cycling among classes, or chronological order. Two collections, three histories, ten nested sizes, and three predetermined divisions give 180 fine-tuning runs. Within a division, the histories differ early but reach the same complete pool. Minority priority is a retrospective label-aware oracle, while chronological ordering describes historical availability rather than future-period evaluation.

\subsection{Selecting Risky Predictions and Comparing Actions}
For every record, three E1 checkpoints each produce ten stochastic dropout predictions \citep{pmlr-v48-gal16}. Their mean probability vector determines E1's label and the low-confidence and predictive-entropy scores; mutual information and variation ratio additionally require the individual stochastic predictions. We rank records by each score, select the riskiest 10\%, and apply every candidate action to the same predictions.

Because these predictions add computation before E2 or the LLM is used, we compare one, three, or five passes per checkpoint with the 30-prediction reference. Fewer passes can change E1's averaged label and risk ordering; AspectBench uses fresh inference, while the external analyses replay stored pass-level predictions.

Low confidence uses E1's chosen-class probability; entropy measures the mean distribution's diffuseness; mutual information measures disagreement among stochastic distributions; and variation ratio counts changes in the winning label. We call the rule converting a ranking into a selected set the \emph{gate}. It chooses which predictions receive a second decision but does not change labels itself.

E1 remains unchanged outside the selected set. Within it, we replace E1 with E2, average their probabilities, choose the more-confident encoder, or ask the LLM to reconsider E1. DSPy optimises and then freezes the LLM instructions \citep{khattab2023dspycompilingdeclarativelanguage}; masked prompts hide the target name and unmasked prompts retain it. A correction repairs a wrong E1 label, while a harm changes a correct one into an error. The matched-action analysis uses 2,000 paired whole-document resamples; policy and quality--compute contrasts use 10,000 paired source-group resamples \citep{efron1993introduction}. Appendix~\ref{sec:detailed-protocol} gives the controls and full protocol.

\subsection{Selecting and Freezing a Complete Policy}

To ask whether development data can identify a useful strategy beforehand, we study SentiCoref and NewsMTSC policies with three choices:
\begingroup
\setlength{\abovedisplayskip}{4pt plus 1pt minus 1pt}
\setlength{\belowdisplayskip}{4pt plus 1pt minus 1pt}
\setlength{\abovedisplayshortskip}{3pt plus 1pt}
\setlength{\belowdisplayshortskip}{4pt plus 1pt minus 1pt}
\begin{equation}
\pi=(s,r,a),
\end{equation}
\endgroup
where $s$ is a risk signal, $r\in\{.05,.10,.20,.30\}$ the processed fraction, and $a$ one of the five encoder or LLM actions above. Signal and fraction select individual predictions, but the dataset-level policy applies one action to all of them. Top-$k$ fixes the fraction within a batch; a numerical threshold permits streaming but its call rate can shift.

Development predictions are out of sample: each record is predicted only by checkpoints that excluded its source document. After aligning E1 and E2 and removing documents used for DSPy, we divide the remainder by source into 70\% discovery and 30\% confirmation. The encoder-input variant was selected earlier by held-out development Macro-F1 and is distinct from masking the later LLM prompt. No test record enters this process, although confirmation remains conditional on that upstream choice.

Discovery selects the best of 80 policies by Macro-F1, breaking ties by lower processing rate and a fixed action-cost order. Confirmation supplies only a point-estimate veto: if that exact policy does not improve on E1, the system makes no additional call. This uncalibrated split, with no predeclared minimum effect, is not formal confirmation. We freeze the survivor, evaluate it on test labels, and only then inspect the fixed grid's hindsight best. Appendix~\ref{sec:policy-transfer-protocol} gives the full construction.

\section{Results}
\subsection{RQ1, Acquire: How the Labelled Collection Grows}
Appendix Figure~\ref{fig:acquisition-full} follows nested histories rather than unrelated samples at isolated sizes. In Slovenian AspectBench, minority priority leads random by 6.27 Macro-F1 points at 1\% and 7.51 points at 5\%, when the rare class can otherwise almost disappear. The pattern is not monotonic: random leads by 5.24 points at 2\% and also leads from 10 through 40\%. Random ordering is therefore a strong default, even though targeted acquisition can be valuable at particular very small sizes.

This reflects two competing needs. Minority priority can rescue a pilot that barely represents a required class, but continuing it delays the broader mix that random ordering supplies. Because several dimensions change together, the experiment neither isolates one cause nor identifies when targeted acquisition should stop. It is also a label-aware oracle: a live campaign must first find likely minority records.

Chronological ordering exposes another coverage risk: at 20\%, HBS reaches 49.46 Macro-F1 versus 75.42 for random. The fixed holdout means this concerns historical coverage, not future generalisation. Identical 100\% pools still vary by 1.36 HBS and 3.32 SL points, so small gaps may reflect training variation. The full curves and screening-cost sensitivity appear in Appendix~\ref{sec:acquire-support}.

\subsection{RQ2, Detect: Which Predictions Are Worth Revisiting?}

The least-confident 10\% contains a much larger share of errors than each complete test set (Table~\ref{tab:detect}). In HBS, the error rate rises from 13.18\% overall to 42.22\% among the selected predictions; in SL it rises from 7.60\% to 33.59\%. If every selected error could be corrected without changing any correct prediction, the resulting Macro-F1 increase would be 5.05--10.36 points.

Perfect correction is an unattainable bound that repairs every selected error without harming a correct prediction; it measures opportunity, not realised model performance.

\setlength{\intextsep}{5pt plus 1pt minus 1pt}
\begin{table}[!h]
\centering
\footnotesize
\setlength{\tabcolsep}{4pt}
\begin{tabular}{@{}lrrrr@{}}
\toprule
Dataset & Base & Selected & Concentration & Perfect \\
 & error & error & & $\Delta$F1 \\
\midrule
HBS & 13.18 & 42.22 & 3.20$\times$ & +5.75 \\
SL & 7.60 & 33.59 & 4.42$\times$ & +10.36 \\
SentiCoref & 20.05 & 41.67 & 2.08$\times$ & +9.14 \\
NewsMTSC & 16.49 & 45.22 & 2.74$\times$ & +5.05 \\
\bottomrule
\end{tabular}
\caption{E1 error rates (\%) overall (Base) and in its least-confident 10\% (Selected); concentration is their ratio. Perfect $\Delta$F1 is the oracle gain from correcting all selected errors without harms.}
\label{tab:detect}
\end{table}

Low confidence, entropy, and mutual information produce similar error rankings (error-AUROC 0.695--0.822); variation ratio is weakest on each collection (0.650--0.749). Calibration measures something different: expected calibration error (ECE) asks whether confidence matches accuracy, whereas error AUROC asks whether errors appear early in the ranking. Appendix~\ref{sec:calibration-ranking} reports both.

Preferred signals vary across some action--rate combinations and development partitions, but reveal no stable action-specific mapping. We therefore select the signal jointly with rate and action rather than claim that each corrective model has a generalisable uncertainty measure. An explanation-derived Gemma score is also unsuitable here as a compute-saving gate because it requires generating the LLM response before selecting records (Appendix~\ref{sec:detect-selection}).

Because uncertainty estimation itself adds compute, we also test fewer stochastic predictions. Nine predictions (three per checkpoint) capture 98.5--104.9\% as many E1 errors as the 30-prediction queue, with 81.4--91.3\% overlap; the largest E1 Macro-F1 change is $-0.188$ on NewsMTSC. This is promising, not proof of equivalence or a minimum; Appendix~\ref{sec:reduced-mc} gives the full sensitivity analysis.

\subsection{RQ3, Resolve: Which Actions Actually Correct Them?}

RQ2 establishes opportunity, not corrective utility. We therefore apply every action to the same least-confident 10\% (Table~\ref{tab:strategy-overview}). On SentiCoref, these 216 predictions include 90 E1 errors. E2 replacement corrects 24 of them but damages 59 predictions that E1 had classified correctly; averaging corrects 20 and harms 42. Masked Gemma corrects only 8 and harms 17, while target-retaining Gemma corrects 11 and harms 9. The second encoder therefore repairs more errors than Gemma but introduces still more mistakes. Gemma repairs fewer errors while preserving substantially more of E1's correct decisions. Identifying an error-rich subset is not enough: the corrective model must also avoid damaging predictions that were already correct. For SentiCoref, perfectly correcting the selected errors could increase Macro-F1 by 9.14 points, yet the best evaluated action on that same queue gains only 0.90 points. Most of the identified correction opportunity therefore remains unrealised.

The masked result also explains why correction counts and Macro-F1 can move differently. All eight corrections concern minority classes (three negative and five positive), whereas 16 of the 17 harms concern neutral records. Positive-class F1 rises 1.009 points, while negative and neutral fall 0.436 and 0.289; equal-class averaging therefore changes Macro-F1 by only $+0.095$ points even though nine net additional errors reduce accuracy by 0.42 points. Its paired Macro-F1 interval includes zero, so this is a metric-composition explanation rather than evidence of reliable benefit.

\setlength{\dbltextfloatsep}{4pt plus 1pt}
\begin{table*}[t]
\centering
\scriptsize
\setlength{\tabcolsep}{2.8pt}
\renewcommand{\arraystretch}{0.9}
\begin{tabular}{@{}llrrrrr@{}}
\toprule
Scope & Dataset & E1 & E2 & E1+E2 average & \shortstack{Gemma without\\E1 information} & \shortstack{E1-assisted\\Gemma} \\
\midrule
\multirow{4}{*}{\shortstack[l]{All\\records}} & AspectBench HBS & 82.63 & +0.33 & +1.22 & $-22.90$ & $-1.13$ \\
 & AspectBench SL & 74.53 & +0.72 & +1.27 & $-31.98$ & $-3.80$ \\
 & SentiCoref SL & 59.91 & $-6.41$ & $-2.92$ & $-10.70$ & +0.40 \\
 & NewsMTSC EN & 82.14 & +0.48 & +1.89 & $-5.46$ & +0.50 \\
\midrule
Scope & Dataset & E2 replacement & E1+E2 average & \shortstack{More-confident\\encoder} & \shortstack{E1-assisted Gemma\\masked / unmasked} & Perfect correction \\
\midrule
\multirow{4}{*}{\shortstack[l]{Selected\\10\%}} & AspectBench HBS & +0.50 & +0.77 & +0.74 & +0.58 / +0.53 & +5.75 \\
 & AspectBench SL & +1.15 & +1.14 & +1.14 & +1.06 / +1.23 & +10.36 \\
 & SentiCoref SL & $-3.30$ & $-2.52$ & $-2.46$ & +0.09 / +0.90 & +9.14 \\
 & NewsMTSC EN & +1.51 & +1.51 & +1.75 & +0.27 / +1.28 & +5.05 \\
\bottomrule
\end{tabular}
\caption{Macro-F1 comparison: E1 is absolute; all other scores are point changes. All-record Gemma retains the target name; E1-assisted Gemma also receives E1's prediction. Selected-10\% actions modify the same queue and retain E1 elsewhere. A slash separates masked and unmasked Gemma; perfect correction is an oracle upper bound.}
\label{tab:strategy-overview}
\end{table*}

The reversal concerns this E1/E2 pairing, not second encoders in general. With 10,000 paired source-group bootstrap samples, SentiCoref averaging changes Macro-F1 by $-2.52$ points $[-4.49,-0.76]$, whereas target-retaining Gemma changes it by $+0.90$ $[-0.28,+2.31]$. On NewsMTSC, both averaging (+1.51) and Gemma (+1.28) improve on E1.

The comparison shows that a model's performance when used on every record can disagree with its value as a selective corrective action, although Gemma's SentiCoref gain remains uncertain. Sending more records is not consistently better either: increasing the selected fraction from 10\% to 20\% helps only seven of 18 dataset--prompt curves. An exploratory analysis also found no reliable improvement from using Gemma's explanations and structured output fields for additional verification; these features are available only after the LLM has been called. Appendix Sections~\ref{sec:resolve-support}--\ref{sec:reasoning-output} provide paired transitions, rate comparisons, and overlapping language/script subsets.

\subsection{RQ4, Select: Can Development Data Choose the Complete Policy?}

Matched-action results reveal what worked after test outcomes are known. RQ4 instead uses development data to choose one of 80 signal--rate--action tuples. Confirmation applies its point-estimate veto to that choice, which is then frozen before test. Test labels evaluate rather than select it, although the study remains retrospective rather than prospectively blind.

Despite Gemma's weaker standalone performance (Table~\ref{tab:strategy-overview}), both development-selected policies use it to reconsider E1 predictions. SentiCoref selects entropy, 30\% routing, and masked Gemma, gaining 1.02 Macro-F1 points $[-0.53,+2.77]$; the interval includes zero. NewsMTSC selects low confidence, 30\%, and target-retaining Gemma, gaining 1.71 points $[+0.48,+2.99]$. These source-group bootstrap intervals condition on fitted models, frozen selections, and stored answers.

Neither policy is the grid's hindsight test optimum. Their \emph{empirical selection regret}, the gap to that optimum, is 3.13 points on SentiCoref: its hindsight best uses low confidence, 30\%, and target-retaining Gemma for +4.15. NewsMTSC's is 0.55: low confidence, 20\%, and the more-confident encoder reaches +2.26. Regret is neither improvement over E1 nor a significance test; Appendix Figure~\ref{fig:policy-transfer} gives the full curves.

\setlength{\textfloatsep}{3pt plus 1pt}
\begin{table}[t]
\centering
\scriptsize
\setlength{\tabcolsep}{1.7pt}
\begin{tabular}{@{}llrcc@{}}
\toprule
Dataset & Action & Route & $\Delta$F1 [95\% CI] & \shortstack{GPU-min\\per 1k} \\
\midrule
SentiCoref & Keep E1 & 0\% & 0.00 [--] & 0.00 \\
 & Avg(E1,E2) & 10\% & $-2.52$ [$-4.49,-0.76$] & 0.03 \\
 & Unmasked Gemma & 10\% & $+0.90$ [$-0.28,+2.31$] & 7.89 \\
 & Selected Gemma & 30\% & $+1.02$ [$-0.53,+2.77$] & 43.71 \\
\addlinespace
NewsMTSC & Keep E1 & 0\% & 0.00 [--] & 0.00 \\
 & Avg(E1,E2) & 10\% & $+1.51$ [$+0.38,+2.68$] & 0.03 \\
 & Unmasked Gemma & 10\% & $+1.28$ [$+0.39,+2.21$] & 4.38 \\
 & Selected Gemma & 30\% & $+1.71$ [$+0.48,+2.99$] & 13.14 \\
\bottomrule
\end{tabular}
\caption{Held-out Macro-F1 changes from E1 with 95\% paired source-group bootstrap CIs (10,000 resamples), and additional allocated warm GPU-minutes per 1,000 records. Selected Gemma uses a masked prompt on SentiCoref; other Gemma rows retain the target. Compute excludes E1 and deployment overhead.}
\label{tab:quality-cost}
\end{table}

Table~\ref{tab:quality-cost} uses 10\% probability averaging as a representative inexpensive encoder action. Retrospectively, choosing NewsMTSC's more-confident encoder gains 1.75 points at approximately the same 0.03 additional GPU-minutes per 1,000 records, but development data did not select it. Appendix~\ref{sec:resources} reports paired estimates.

Cost differences are substantial, while additional gains are less clear. SentiCoref's selected policy beats averaging by 3.54 points $[+0.86,+6.25]$, but beats 10\% target-retaining Gemma by only 0.12 $[-1.34,+1.70]$; on NewsMTSC, its advantage over averaging is 0.20 $[-1.19,+1.61]$. Encoder estimates use warm throughput and one XLM-R proxy; Gemma uses historical job times and allocated capacity. They measure additional compute, not complete deployment expenditure.

A secondary experiment chooses an action per selected record rather than one action per dataset. These exploratory correctness routers do not consistently beat fixed actions and some require extra information; Appendix~\ref{sec:contextual-routing} reports them. This does not rule out routers trained for incremental utility and cost.

With each policy fixed, retaining only 25\% of source groups is harmful in 26.7\% of SentiCoref and 6.7\% of NewsMTSC resamples. These reuse existing tests, not future deployments. Appendix~\ref{sec:policy-robustness} separates sample-size and class-mixture effects from threshold call-rate drift.

\section{Discussion, Conclusion, and Future Work}

Targeted acquisition can help when a required class is scarcely represented, but its non-monotonic advantage provides no universal stopping point. The evidence instead supports checking early class support before broadening coverage; acquisition remains experimentally separate from selective correction.

After training, uncertainty identifies predictions worth checking but cannot determine what another model will do. On SentiCoref, an oracle that corrected every error in the selected 10\% without harming correct predictions would improve Macro-F1 by 9.14 points. The best evaluated corrective action on that same queue gains only 0.90 points. The second encoder repairs more errors than Gemma but also overturns many more correct decisions. This gap between correction opportunity and realised improvement shows why corrective models must be evaluated on the predictions they actually receive.

Development selection chose E1-assisted Gemma on both external datasets despite its weaker standalone performance, using different risk signals and prompt masking. Both policies have positive test point estimates, but SentiCoref's interval includes zero and neither policy matches the hindsight optimum. This shows why standalone model rankings alone cannot determine the preferred corrective action, without establishing that the choices will generalise to new datasets.

Computation changes the practical interpretation: inexpensive encoder actions approach the selected Gemma gain on NewsMTSC, whereas encoder averaging harms SentiCoref and Gemma's advantage remains uncertain. These recorded and proxy estimates compare evaluated options, not a common economic objective. For deployment, this suggests comparing corrective models on the predictions they would actually receive, alongside inexpensive alternatives and the additional computation each requires.

Future work should test controlled South Slavic data combinations, cross-dataset routers, and complete policies on later periods and domains. Utility- and cost-aware routers could target incremental benefit directly. Joint annotation and inference allocation, however, requires a study with a common economic objective.

\captionsetup[table]{skip=10pt}
\setlength{\textfloatsep}{20pt plus 2pt minus 4pt}
\setlength{\dbltextfloatsep}{20pt plus 2pt minus 4pt}
\setlength{\intextsep}{12pt plus 2pt minus 2pt}
\section{Limitations}

The acquisition study is retrospective. Minority-prioritised ordering uses known labels, so it omits the retrieval errors and screening effort that a live campaign would incur. Chronological acquisition processes oldest records first but evaluates on the same reserved test set; it is not a future-period test. Each policy, data fraction, and predetermined data division has one training run, and the identical-data result shows that training variation can be large enough to obscure small policy differences. We also do not train and evaluate every downstream gate and corrective action along every acquisition history. Acquire and selective escalation are linked planning decisions, but this study does not estimate a causal path from an acquisition policy to later escalation quality.

The external datasets do not provide a controlled language comparison. SentiCoref requires a three-class conversion and a custom test partition that keeps all records from one document together, whereas NewsMTSC uses its official test and substantially shorter contexts. The reversal between them demonstrates that the best additional action does not transfer automatically. It does not identify language, document length, or dataset construction as the cause. In particular, many SentiCoref documents are long, so an additional encoder designed to read longer text is a useful follow-up. The current result concerns one XLM-R model, not second encoders in general.

The complete development-to-test policy-selection experiment has only two independent external replications. Test labels do not select the reported tuple, but this is still a retrospective study rather than a prospectively blind evaluation because the datasets and some test outcomes were already familiar during the wider research process. The 80-policy grid can overfit the finite discovery set. Target-masked versus target-retaining encoder-input variants were selected on held-out development Macro-F1 before the 70/30 policy split, so there is no test leakage but the confirmation partition is conditional on that upstream modelling choice. The split was not power-calibrated, and the smaller ``confirmation'' partition only vetoes a winner whose point-estimate gain is not positive; no minimum useful effect or uncertainty criterion was predeclared. It neither provides formal statistical confirmation nor validates the ranking of every candidate. The source-group bootstrap intervals condition on the frozen predictions and do not include retraining, policy-search, prompt-selection, or future-shift uncertainty; there is no sampling distribution for selection regret. Neither chosen tuple exactly matches the hindsight optimum. AspectBench provides out-of-fold contextual-router predictions for seven overlapping language or script subsets, not another complete signal--rate--action selection experiment.

Only one principal type of E2 model is evaluated per external dataset. The LLM results concern Gemma 3 27B under instruction sets optimised with DSPy and fixed before test evaluation and one response per item. The uncertainty intervals condition on fitted encoders and LLM outputs; they do not include variation from training the encoders again, searching for different instructions, or generating another LLM response. Selecting the least-confident 10\% is a fixed comparison budget rather than a universal deployment threshold. The full uncertainty estimate uses three E1 checkpoints with ten Monte Carlo dropout predictions each, giving 30,000 primary-encoder forwards per 1,000 records before a 10\% queue is chosen; the nine-prediction reconstruction reduces this to 9,000. Although all four principal collections now support reduced-pass analysis, nine predictions were examined without a predeclared acceptable-loss margin and do not establish equivalence, universal sufficiency, or optimality. AspectBench corrective-action rows keep stored E2 decisions fixed while changing the queue, and cached Gemma answers contain historical E1 evidence; neither is a newly executed end-to-end nine-prediction pipeline. A deterministic three-checkpoint ensemble is a cheaper baseline that remains competitive in error concentration, although it reproduces less of the 30-prediction AspectBench queue.

Repeated dropout predictions, E2 calls, and LLM calls have different response-time, memory, and energy costs. The operational ledger combines direct measurements, reconstructed outcomes, historical timings, and assumed prices rather than observing one end-to-end deployment. Its labour rates, document-to-target ratio, screening yield, hardware allocation, utilisation, exchange rate, traffic, and retraining horizon are replaceable scenario inputs. The acquisition sizes and corrective actions were not crossed prospectively under equal monetary budgets, so the study cannot determine whether the next euro should buy another label or another call. The small direct GPU-rental amounts also exclude engineering, unsuccessful experiments, quality assurance, infrastructure, maintenance, storage, and staff time.

AspectBench language and script subsets reveal meaningful heterogeneity but overlap and cannot be treated as independent replications. They also do not isolate pooling from training-set size, preprocessing, data distribution, or model selection. A controlled pooled-versus-separate comparison would require matched architectures, preprocessing, seeds, training-data quantities, and target-language tests.

The contextual router is deliberately modest: separate class-balanced logistic models predict each action's correctness inside a fixed 30\% low-confidence region, then a record receives the action with the highest estimated correctness. It optimises a decomposable correctness target rather than Macro-F1, and the E1+E2 version must already run E2 before routing. Its oracle comparison uses gold labels and is unattainable. The results therefore test whether this particular learned action rule improves on constant actions; they do not establish that contextual routing cannot work. A stronger study would learn expected incremental utility and cost across the full stream, compare pooled and dataset-specific routers on more independent datasets, and include human review.

The robustness analysis resamples existing test documents. Repeated subsets measure sensitivity to sample size, while reweighting changes class prevalence but does not create new language, topics, events, annotation rules, or relationships between text and sentiment. The 30 seeds are not independent deployments. Only NewsMTSC retains two sufficiently large source groups, and neither external dataset has reliable timestamps for a chronological evaluation. A genuinely prospective later-period or new-domain test remains necessary. A fixed risk-score threshold can also change the fraction routed when the data distribution changes; a larger gain obtained with more calls is not a same-budget improvement.

Prospective annotation studies should measure the cost of finding likely examples from rare classes, while a strictly later test period would separate acquisition efficiency from changes in language and sentiment over time. A deployment study can then combine accuracy, the importance assigned to each sentiment class, response time, and cost in one objective chosen before evaluation.

\section{Ethical Considerations}

Entity-level sentiment predictions about named people and organisations can affect reputation and may be used in monitoring workflows. The system should therefore support analyst prioritisation rather than silently turn uncertain model outputs into consequential judgements. Its class imbalance and uneven language/script coverage may produce systematically different error rates for minority sentiments or underrepresented groups, and the present subgroup results do not establish fairness across entities, sources, or demographic categories. Deployment should retain provenance to the source article, expose uncertainty and escalation decisions, permit human correction, and monitor class- and subgroup-level errors over time.

\ifdefined\NAMEDVERSION
\section*{Acknowledgments}
Generative AI tools, including OpenAI ChatGPT and Codex (models in the GPT-5.6 and GPT-6 families) \citep{openai2026gpt56,openai2026gpt6} and Google Gemini 3.1 Pro and 3.8 Flash \citep{google2026gemini31pro,google2026gemini38flash}, assisted with language revision, literature discovery, structural critique, consistency checking, and code development and automation during manuscript preparation. Their outputs were not treated as scientific evidence. The authors checked cited sources and experimental results, reviewed and edited the generated material, and take full responsibility for the manuscript.
\fi

\bibliography{custom}

\clearpage
\appendix
\raggedbottom

\section{Detailed Methods and Supporting Results}
\label{sec:all-strategies}

This appendix provides the configurations, checks, and expanded comparisons needed to interpret the main results. We first describe the protocol shared across analyses and distinguish measured computation from reconstructed or assumed quantities. The remaining sections follow the question each analysis supports, as summarised below.

\begin{center}
\centering
\footnotesize
\setlength{\tabcolsep}{3pt}
\begin{tabularx}{\columnwidth}{@{}lX@{}}
\toprule
Main question & Supporting appendix evidence \\
\midrule
RQ1: Acquire & Dataset composition, complete acquisition curves, and class-extension checks (Section~\ref{sec:acquire-support}) \\
RQ2: Detect & Reduced-pass sensitivity (Section~\ref{sec:reduced-mc}) and calibration, ranking, and selection rules (Section~\ref{sec:calibration-ranking}) \\
RQ3: Resolve & Complete strategy matrix and matched actions (Section~\ref{sec:resolve-support}), routing rates (Section~\ref{sec:gate-rate}), encoder combinations (Section~\ref{sec:encoder-language}), and LLM-output checks (Section~\ref{sec:reasoning-output}) \\
RQ4: Select & Development-to-test generalisation, per-record routing, and population sensitivity (Sections~\ref{sec:policy-transfer-results}--\ref{sec:policy-robustness}) \\
Resources & Measured runtime and illustrative cost accounting (Section~\ref{sec:resources}) \\
\bottomrule
\end{tabularx}
\captionof{table}{Where the appendix supports each main-text research question.}
\label{tab:appendix-map}
\end{center}

Not every appendix analysis plays the same evidential role. Some expand the principal comparison, some reconstruct an alternative configuration from stored outputs, and others diagnose heterogeneity or failure modes. Table~\ref{tab:evidence-status} defines these roles once; they describe how we use the evidence, not a preregistration status. The study as a whole is retrospective.

\begin{table*}[t]
\centering
\footnotesize
\setlength{\tabcolsep}{4pt}
\begin{tabularx}{\textwidth}{@{}p{.19\textwidth} p{.29\textwidth} X@{}}
\toprule
Evidence role & Purpose & Analyses \\
\midrule
Main comparisons & Directly support the four headline answers under their stated retrospective protocols. & Nested acquisition histories; 30-prediction detection at the fixed 10\% comparison budget; corrective actions on identical queues; policies selected on development data and frozen for held-out evaluation. \\
Reconstructed sensitivities & Reuse cached predictions, model answers, timings, or scenario inputs to ask how a measured result changes under another configuration. & Reduced-pass queue reconstruction; stored-output routing-rate curves; throughput extrapolations and illustrative deployment accounting. \\
Exploratory diagnostics & Expose heterogeneity, possible mechanisms, or failure modes without serving as independent confirmation of the headline result. & Extended routing-rate sensitivities; overlapping language and script subsets; contextual routers; structured-output checks; resampled sample-size and class-mixture analyses. \\
\bottomrule
\end{tabularx}
\caption{Evidence roles in the appendix. ``Main'' identifies the comparisons used for the paper's principal answers; it does not mean prospectively registered. Reconstructed and exploratory results retain the qualifications given in their sections.}
\label{tab:evidence-status}
\end{table*}

\subsection{How the Research Questions Relate}
\label{sec:operational}

\begin{figure*}[!t]
\centering
\resizebox{.98\textwidth}{!}{%
\begin{tikzpicture}[
  every node/.style={font=\footnotesize,align=center},
  box/.style={draw,line width=.45pt,rounded corners,fill=blue!6,text width=31mm,minimum height=13mm,inner sep=1.5mm},
  decision/.style={draw,line width=.45pt,rounded corners,text width=35mm,minimum height=19mm,inner sep=1.5mm},
  outcome/.style={draw,line width=.45pt,rounded corners,fill=cyan!7,text width=33mm,minimum height=13mm,inner sep=1.5mm},
  arrow/.style={-{Latex[length=2mm]},line width=.65pt}
]
\node[box] (prior) at (0,0) {\citet{chatterjee2026evaluating}: AspectBench modelling, learning curves, and selective deferral};
\node[decision,fill=green!7] (acquire) at (4.5,1.45) {\textbf{RQ1: Acquire}\\Compare nested orderings of already labelled AspectBench records};
\node[outcome] (acqresult) at (9.0,1.45) {Acquisition result; histories end at the same pool};
\node[decision,fill=purple!7] (detect) at (4.5,-1.45) {\textbf{RQ2: Detect}\\With trained systems fixed, compare risk signals and queues};
\node[decision,fill=orange!13] (resolve) at (9.0,-1.45) {\textbf{RQ3: Resolve}\\Apply competing actions to the same selected records};
\node[decision,fill=yellow!18] (select) at (13.5,-1.45) {\textbf{RQ4: Select}\\Choose $(s,r,a)$ on development data and freeze it before test};
\node[outcome] (measure) at (18.0,-1.45) {Held-out gain, regret, contextual routing, and sensitivity};
\node[font=\footnotesize\itshape,text width=50mm] at (13.5,1.45) {RQ1 informs the same practical setting, but its acquisition histories are not propagated through the later escalation experiments.};
\draw[arrow] (prior) -- (acquire);
\draw[arrow] (acquire) -- (acqresult);
\draw[arrow] (prior) -- (detect);
\draw[arrow] (detect) -- (resolve);
\draw[arrow] (resolve) -- (select);
\draw[arrow] (select) -- (measure);
\end{tikzpicture}%
}
\caption{Experimental relationships among the research questions. RQ1 independently reorders already labelled AspectBench records and does not feed each acquisition history into the later experiments. RQ2--RQ4 form the stronger methodological sequence: risk signals define a queue, actions are compared on matched queues, and development data select a complete policy before test.}
\label{fig:rq-structure}
\end{figure*}
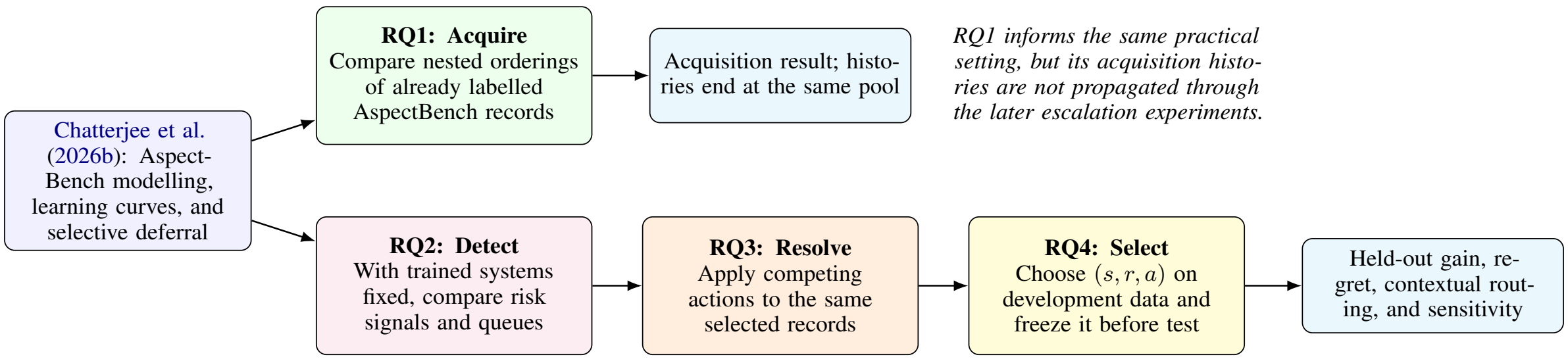

The two branches answer different experimental questions. RQ1 asks how an already labelled AspectBench pool might have grown; we do not retrain the complete selective-correction system along every acquisition history. RQ2 instead begins with trained systems and ranks predictions by estimated risk. RQ3 asks what competing models do to those same predictions, and RQ4 asks whether development data can choose the ranking, fraction, and action before test evaluation.

This progression separates four forms of evidence: whether the collection covers required classes and contexts, whether a risk score places errors early, whether an action repairs more predictions than it harms, and whether the complete choice survives held-out evaluation. The associated computation is also unlike. Repeated E1 predictions, E2 calls, and autoregressive Gemma calls should not be collapsed into one runtime number. We therefore define those units in the shared protocol and resource sections before reporting question-specific results.

\subsection{Detailed Experimental Protocol}
\label{sec:detailed-protocol}

Each evaluation supplies an article and a target entity. We score negative, neutral, or positive document-level ELSA predictions; target extraction and joint target--sentiment extraction are outside the study.

AspectBench is an already labelled industrial archive. Trained analysts assigned sentiment to specified document--entity pairs in the production workflow, after which the published study cleaned and restructured the records for modelling \citep{chatterjee2026evaluating}. The present acquisition experiment changes only the order in which those known records become available. SentiCoref contains Slovenian news, named entities, coreference chains, and five-level sentiment annotations \citep{zitnik2019senticoref,zitnik2022target}. We map these to negative, neutral, and positive, reserve 15\% of source documents for testing, and keep every record from a document in one partition. NewsMTSC uses its official real-world English political-news test collection \citep{hamborg-donnay-2021-newsmtsc}.

The corrective-action pairs are BERTi\'{c}/XLM-R for AspectBench HBS, SloBERTa/XLM-R for AspectBench SL and SentiCoref, and XLM-R/mDeBERTa-v3 for NewsMTSC \citep{ljubesic2021bertictransformerlanguage,ulcar2021sloberta,conneau2020unsupervisedcrosslingualrepresentationlearning,he-etal-2023-debertav3}. The first model in each pair is E1 and the second is E2. Macro-F1 computes F1 separately for negative, neutral, and positive and weights the three classes equally. QWK additionally recognises their order and penalises a negative--positive confusion more heavily than a negative--neutral or neutral--positive confusion \citep{cohen1968weighted}. Accuracy lets us express a selective action directly: for selected record $i$, E1 prediction $e_i$, action prediction $a_i$, and reference label $y_i$, a correction has $e_i\ne y_i$ and $a_i=y_i$, while a harm has $e_i=y_i$ and $a_i\ne y_i$. Hence
\begin{equation}
\Delta\mathrm{Acc}=\frac{\#\mathrm{corrections}-\#\mathrm{harms}}{N}.
\end{equation}
Macro-F1 also depends on which classes change, so counts and final class-balanced scores are both required.

To select predictions for reconsideration, the primary rule orders records by the maximum averaged E1 probability and takes the least-confident 10\%, using record ID to break ties reproducibly. Predictive entropy instead measures the spread of the averaged probabilities, mutual information the disagreement among stochastic predictions, and variation ratio the frequency with which the modal hard label is not selected. We refer to such a selection rule as a gate only after this ordering has been defined. In reduced-compute trials, every record and checkpoint uses the same pass-position subset. This prevents item-wise test selection and corresponds to a configuration that could be fixed before use.

For the corrective-action comparison, E1 retains every prediction outside the selected set. Within it, the candidate action may replace E1 with E2, average the E1 and E2 probabilities, choose the more confident encoder, or invoke E1-assisted Gemma 3 27B. Gemma receives the article, target, E1 label, class probabilities, and uncertainty scores. Masked programs replace target mentions with a neutral marker, whereas unmasked programs retain the name. DSPy optimises the instructions on low-confidence validation records, after which they are frozen for test evaluation. Controls also give this context to Gemma for every record, or apply E2, standalone Gemma, or encoder averaging throughout the test set.

The principal matched-action figure uses 2,000 paired whole-source-document resamples. The quality--compute table separately uses 10,000 paired source-group resamples: every sampled source contributes all of its target records, and every strategy is scored on the same resample. Both analyses condition on fitted encoders and stored LLM responses. Their intervals therefore exclude retraining, policy search, prompt selection, another generation, and future distribution shift.

\subsection{Reproducibility and Resource Accounting}
\label{sec:resources}

Computation has no single natural unit across these experiments. Training an encoder on one GPU, serving concurrent LLM requests, and repeating E1 thirty times all use accelerators but describe different workloads. We therefore keep the training procedure, serving configuration, pass count, device time, and concurrent wall time separate. \emph{Measured} quantities come from recorded jobs or the A40 benchmark; \emph{reconstructed} quantities are calculated from stored predictions without retraining; and \emph{assumed} quantities belong to a replaceable accounting scenario. Table~\ref{tab:encoder-protocol} gives the encoder settings. The acquisition grid does not use Monte Carlo dropout. The external action comparison combines ten dropout predictions from each of three trained models, yielding thirty probability vectors per item.

\subsubsection{Annotation Units and Illustrative Cost Accounting}
\label{sec:cost-accounting}

An annotation budget must distinguish the text that an annotator reads from the target for which a label is recorded. The dataset counts in this paragraph are measured; the labour and article-volume scenario that follows is assumed. In the training and validation pools, AspectBench HBS contains 70,307 article--entity records from 60,496 source articles (1.16 records/article), and AspectBench SL contains 43,863 from 35,231 (1.25). The external collections have different structures: NewsMTSC contains 8,733 records from 7,733 articles (1.13), whereas SentiCoref contains 12,211 from 707 documents (17.27). The illustrative 7,000 articles per 10,000 labels below is therefore an assumption, not a measured ratio that can be transferred among these datasets.

The scenario assumes 40 expert-hours for guidelines and pilot work, 16 annotator-hours of training, eight expert-hours of supervision, annotator labour at \texteuro25/hour, expert labour at \texteuro100/hour, 2.5 minutes to read a new article, one minute to label a target, and two minutes of expert review for 10\% of targets. These values are editable inputs, informed by rather than measured from the present project; annotation costs and times vary with task and sampling policy \citep{haertel-etal-2008-assessing}, and training and supervision should not be hidden inside a per-label rate \citep{schepers-etal-2025-price}.

\begin{table}[!tbp]
\centering
\footnotesize
\setlength{\tabcolsep}{4pt}
\resizebox{\columnwidth}{!}{%
\begin{tabular}{@{}llr@{}}
\toprule
Scenario component & Basis & Cost (\texteuro) \\
\midrule
Setup & Guidelines, training, supervision & 5,200.00 \\
Production and review & 10,000 labels; 7,000 articles & 14,791.67 \\
E1 and E2 training & Three checkpoints each & 0.60 \\
DSPy calibration & Allocated endpoint time & 4.40 \\
Inference & 100,000 future decisions & 6.90 \\
\midrule
Illustrative total & Mixed experimental workloads & 20,003.56 \\
\bottomrule
\end{tabular}
}
\caption{Illustrative accounting scenario, not an observed AspectBench annotation bill or a single end-to-end deployment. Direct accelerator rental excludes engineering, experimentation, unsuccessful runs, quality assurance, infrastructure, storage, maintenance, and staff time.}
\label{tab:illustrative-cost}
\end{table}

The GPU scenario converts US dollars at an assumed rate of \$1=\texteuro0.92 and uses an October 2026 snapshot of \$0.59 per A40 or RTX A6000 GPU-hour from \href{https://www.runpod.io/pricing}{RunPod pricing (October 2026)}. The three SentiCoref E1 and E2 training workloads occupy 0.509 and 0.596 single-GPU hours, respectively, and cost \texteuro0.276 and \texteuro0.324 at that rate. Those recorded workloads include training, export, deterministic prediction, and ten-pass validation/test prediction; they are lower bounds on obtaining a classifier, not full development costs. The \texteuro4.40 calibration entry is a scenario allocation based on four GPUs for the observed DSPy wall time. Actual endpoint utilisation was not measured.

Minority acquisition introduces a separate screening term. Under random screening at 0.5 minute per candidate and \texteuro25/hour, obtaining 1,000 minority examples costs approximately \texteuro20,833 at 1\% prevalence (100,000 candidates) or \texteuro2,778 at 7.5\% prevalence (13,333 candidates), before final labelling and review. These values are sensitivity calculations, not a retrieval experiment. A learned prescreener would have development cost, imperfect precision, and false negatives; the result consequently supports costing targeted acquisition, not maintaining an artificially balanced corpus indefinitely.

The deployment horizon determines how the upfront total should be interpreted. Holding the scenario's \texteuro19,996.66 setup-through-calibration component fixed, its amortised contribution is approximately \texteuro2.00 per decision over 10,000 decisions, \texteuro0.20 over 100,000, \texteuro0.02 over one million, and \texteuro0.002 over ten million. Marginal gated inference in this scenario is approximately \texteuro0.000069 per decision. These are arithmetic consequences of the assumptions, not measured returns on either labels or calls. Retraining more often shortens the life of an acquired label; higher traffic increases the number of one-use escalation calls.

\begin{table*}[!tbp]
\centering
\scriptsize
\setlength{\tabcolsep}{3.6pt}
\begin{tabular}{@{}lllrrrrll@{}}
\toprule
Workload & Model & GPU & Epochs & Train/eval batch & Accum. & Max. length & Precision & Uncertainty \\
\midrule
Acquisition SL & SloBERTa & H100 & $\leq30$ & 16 / 32 & 2 & 512 & fp32 & none \\
Acquisition HBS & BERTi\'{c} & H100 & $\leq30$ & 16 / 32 & 2 & 512 & fp32 & none \\
SentiCoref E1 & SloBERTa & A40 & $\leq5$ & 8 / 24 & 4 & 512 & bf16 & 30 passes \\
SentiCoref E2 & XLM-R & A40 & $\leq5$ & 8 / 24 & 4 & 512 & bf16 & 30 passes \\
NewsMTSC E1 & XLM-R & A6000 & $\leq5$ & 8 / 24 & 4 & 512 & bf16 & 30 passes \\
NewsMTSC E2 & mDeBERTa-v3 & A6000 & $\leq5$ & 8 / 24 & 4 & 512 & bf16 & 30 passes \\
\bottomrule
\end{tabular}
\caption{Encoder training and uncertainty procedure. An epoch is one complete pass through the training data; batch sizes are the numbers of records processed together during training and evaluation; accumulation is the number of training batches combined before a model update; and maximum length is the token limit. ``fp32'' and ``bf16'' are 32-bit and 16-bit numerical formats. Acquisition runs stop after three evaluations without improvement. External runs stop after two, use an update size of $2\times10^{-5}$, and apply weight decay 0.01 as regularisation. A separately verified update size is unavailable for the acquisition runs, so we do not infer one here.}
\label{tab:encoder-protocol}
\end{table*}

Table~\ref{tab:llm-protocol} describes the specific LLM procedure we evaluate, rather than a generic API call. We optimise the instructions once on validation records, fix the resulting instruction set before test evaluation, and then generate one response for each selected item and instruction set. Each set containing the least-confident 10, 20, 30, 40, or 50\% contains every smaller set. We request answers for the largest set once and reuse each stored answer for every smaller selection rate, so a record is not regenerated merely because it appears in several curves. Retries respond only to transport or API failure; they are not extra alternative answers.

\begin{table}[!t]
\centering
\scriptsize
\begin{tabularx}{\columnwidth}{@{}p{0.42\columnwidth}X@{}}
\toprule
Component & Setting used \\
\midrule
Model producing test answers & Gemma 3 27B, 4-bit quantised \\
Model helping optimise instructions & Qwen 2.5 72B, 4-bit quantised \\
Instruction optimisation & DSPy MIPROv2, medium setting \\
Concurrent optimisation processes & 8 \\
Maximum output & 1,024 tokens \\
Maximum article input & 10,000 characters \\
Serving & vLLM, 24 concurrent requests per server \\
Generation & one response per item/instruction set \\
Retries & transport or API failures only \\
\bottomrule
\end{tabularx}
\caption{LLM instruction-optimisation and answer-generation configuration.}
\label{tab:llm-protocol}
\end{table}

Both language models use 4-bit quantisation, which stores each model parameter with four bits to reduce memory use. A token is a word or word fragment counted by the model; the output limit therefore bounds the length of each answer.

Table~\ref{tab:resource-use} records two kinds of workload. Summed encoder-training durations approximate device time because each run occupies one GPU. LLM jobs share several model servers and issue concurrent requests, so their wall times describe completion under that serving layout. Adding those wall times to the encoder rows as if all were GPU-hours would be misleading.

\begin{table*}[!t]
\centering
\scriptsize
\setlength{\tabcolsep}{4pt}
\begin{tabular}{@{}llllr@{}}
\toprule
Study component & Dataset and runs & Hardware & Measured workload & Time \\
\midrule
RQ1 acquisition & HBS, 90 cells & H100 & summed single-GPU time & 90.08 h \\
RQ1 acquisition & SL, 90 cells & H100 & summed single-GPU time & 49.30 h \\
External E1 & NewsMTSC, 3 trained models & A6000 & training and prediction & 1.16 h \\
External E2 & NewsMTSC, 3 trained models & A6000 & training and prediction & 1.12 h \\
External E1 & SentiCoref, 3 trained models & A40 & training and prediction & 0.51 h \\
External E2 & SentiCoref, 3 trained models & A40 & training and prediction & 0.60 h \\
Gated Gemma deferral & NewsMTSC, 344 records & multiple servers & 30\% selection, generation time & 4.17 min \\
E1+Gemma every record & NewsMTSC, 1,146 records & multiple servers & full-test generation time & 13.07 min \\
Gemma without E1 & NewsMTSC, 1,146 records & multiple servers & full-test generation time & 4.25 min \\
Gated Gemma deferral & SentiCoref, 647 records & multiple servers & 30\% selection, generation time & 18.17 min \\
E1+Gemma every record & SentiCoref, 2,155 records & multiple servers & full-test generation time & 47.86 min \\
Gemma without E1 & SentiCoref, 2,155 records & multiple servers & full-test generation time & 21.40 min \\
\bottomrule
\end{tabular}
\caption{Recorded resource use. Encoder hours add together the time for which each GPU was occupied. LLM rows average the runs that hide or retain the target name while several requests are processed concurrently. They exclude server loading and one-time instruction optimisation, and should not be read as the time for one record or as physical GPU-hours.}
\label{tab:resource-use}
\end{table*}

The 139.39 H100 GPU-hours used for RQ1 are the computational cost of evaluating 180 research cells, not the cost of one deployed acquisition policy. Likewise, the LLM wall times depend on the number of active model servers, the number of simultaneous requests, output length, and whether Gemma processes selected cases or every record. We benchmark E1 warm inference separately below. We did not measure end-to-end energy use or the utilisation of every physical GPU in the historical LLM jobs, so those quantities remain outside the measured ledger.

\subsubsection{Recurring Corrective-Action Compute}

Main-text Table~\ref{tab:quality-cost} reports recurring action compute additional to E1. The 10\% encoder average applies the measured three-checkpoint, three-pass XLM-R A40 rate only to selected records, giving 0.03 allocated GPU-minutes per 1,000 incoming decisions. This is also a proxy for NewsMTSC's mDeBERTa-v3 E2 because no matched warm benchmark was retained. Gemma estimates use observed query-job wall time on two concurrent two-GPU endpoints and multiply by the four allocated GPUs. They do not trace utilisation, per-request latency, loading, idle reservation, orchestration, queueing, storage, or engineering. SentiCoref's 10\% target-retaining and development-selected 30\% masked programs have different observed per-call runtimes, so their difference cannot be attributed to routing rate alone.

As a retrospective sensitivity check, selecting the more confident encoder on the same least-confident 10\% NewsMTSC queue gains 1.75 Macro-F1 points over E1 (95\% paired source-group bootstrap interval $[+0.61,+2.95]$), compared with 1.51 $[+0.38,+2.68]$ for probability averaging. Both require approximately 0.03 additional allocated GPU-minutes per 1,000 incoming records because E2 is evaluated on the same queue. The paired difference between confidence selection and averaging is $+0.25$ points $[-0.19,+0.72]$, while its difference from the development-selected 30\% Gemma policy is $+0.04$ $[-1.33,+1.37]$; neither difference is resolved by these intervals. Although confidence selection was an evaluated action, identifying it as the stronger cheap comparator follows test inspection, so it is not the development-selected RQ4 policy.

One-time resources remain separate from those recurring values. The three-checkpoint E2 workloads use 0.596 allocated GPU-hours for SentiCoref and 1.117 for NewsMTSC. The retained DSPy calibration allocations use 7.204 GPU-hours for SentiCoref's target-retaining program, 8.992 for its selected masked program, and 3.864 for each NewsMTSC Gemma row. These figures describe recorded or allocated workloads, not the full cost of obtaining a classifier or prompt: research search, failed runs, software work, quality assurance, and staff time are excluded.

\subsubsection{Measured A40 Encoder Throughput}

We benchmarked three real masked AspectBench Slovenian SloBERTa checkpoints resident on one NVIDIA A40. The input comprised 1,024 selected test records, with batch size 64 and maximum length 512. Tokenisation was completed before the timed loop, and the reported warm rate is the mean of three repeats using FP16 autocast. Table~\ref{tab:a40-throughput} reports the direct measurement and a linear extrapolation to 100,000 records.

\begin{table}[!tbp]
\centering
\footnotesize
\setlength{\tabcolsep}{4pt}
\begin{tabular}{@{}lrr@{}}
\toprule
E1 configuration & Records/s & A40-h/100k \\
\midrule
3 checkpoints $\times$ 3 MC passes & 56.36 & 0.493 \\
3 checkpoints $\times$ 10 MC passes & 16.77 & 1.656 \\
\bottomrule
\end{tabular}
\caption{Measured warm encoder throughput and extrapolated A40-hours. The 100,000-record column is computed from the 1,024-record benchmark rather than measured in a separate deployment. Nine stochastic predictions means three trained checkpoints with three stochastic forward passes each, not nine independently trained models.}
\label{tab:a40-throughput}
\end{table}

The nine-pass configuration is 3.36 times faster, a 70.2\% reduction in measured E1 forward-pass time. This result is specific to the encoder, hardware, data profile, batch size, and serving procedure above. Preprocessing, cold loading (56.8 seconds in this run), server orchestration, model scheduling, queueing, and any E2 or LLM action are outside the warm loop. Accordingly, the result neither measures the full media-monitoring pipeline nor implies a universal 70\% reduction in operating cost.

The new complete pass exports provide a second, model-specific observation. On one A40, exporting 30 stochastic plus three deterministic predictions for every test record takes 825.0 seconds for 12,407 HBS/BERTi\'{c} records and 533.1 seconds for 7,740 SL/SloBERTa records. These measured jobs include on-the-fly tokenisation but exclude model loading, corresponding to 496.3 and 479.1 forward-records/s. Reconstructing nine- and 30-prediction service time from those rates gives 0.504 versus 1.679 A40-hours per 100,000 HBS decisions and 0.522 versus 1.739 for SL. At the illustrative \$0.59 rate, those busy-compute estimates are approximately \$0.30--\$0.31 for nine predictions and \$0.99--\$1.03 for 30. The complete nine- and 30-prediction services were not separately timed, and these estimates exclude loading, idle reservation, orchestration, storage, and engineering.

For scale, the deployment scenario combines the extrapolated 0.493 A40-hours for nine-pass E1 over 100,000 records with a historical Gemma job that completed 7,740 calls in 17,027.824 seconds. The extrapolation uses the measured 1,024-record warm-inference rate reported above. Dividing the Gemma aggregates gives an effective 2.200 seconds per completed call under concurrent serving; it is not a measured single-request latency. Assuming two A40s for Gemma and a 10\% gate gives approximately 1,636 completed Gemma calls and 16,364 incoming decisions per hour, or 6.11 hours for 10,000 calls. Adding busy E1 and Gemma computation gives 12.715 A40-hours, or about \$7.50 at \$0.59/GPU-hour. Reserving an assumed four-A40 layout continuously for the same 6.11-hour interval gives \$14.42. Both amounts are projections: they exclude cold-start time, batching and scheduling losses, queueing, monitoring, storage, and other infrastructure overhead, and they are not evidence that a four-A40 pipeline was prospectively deployed.

\subsection{RQ1 Support: Dataset and Acquisition Details}
\label{sec:acquire-support}

\begin{figure*}[!t]
\centering
\includegraphics[width=.98\textwidth]{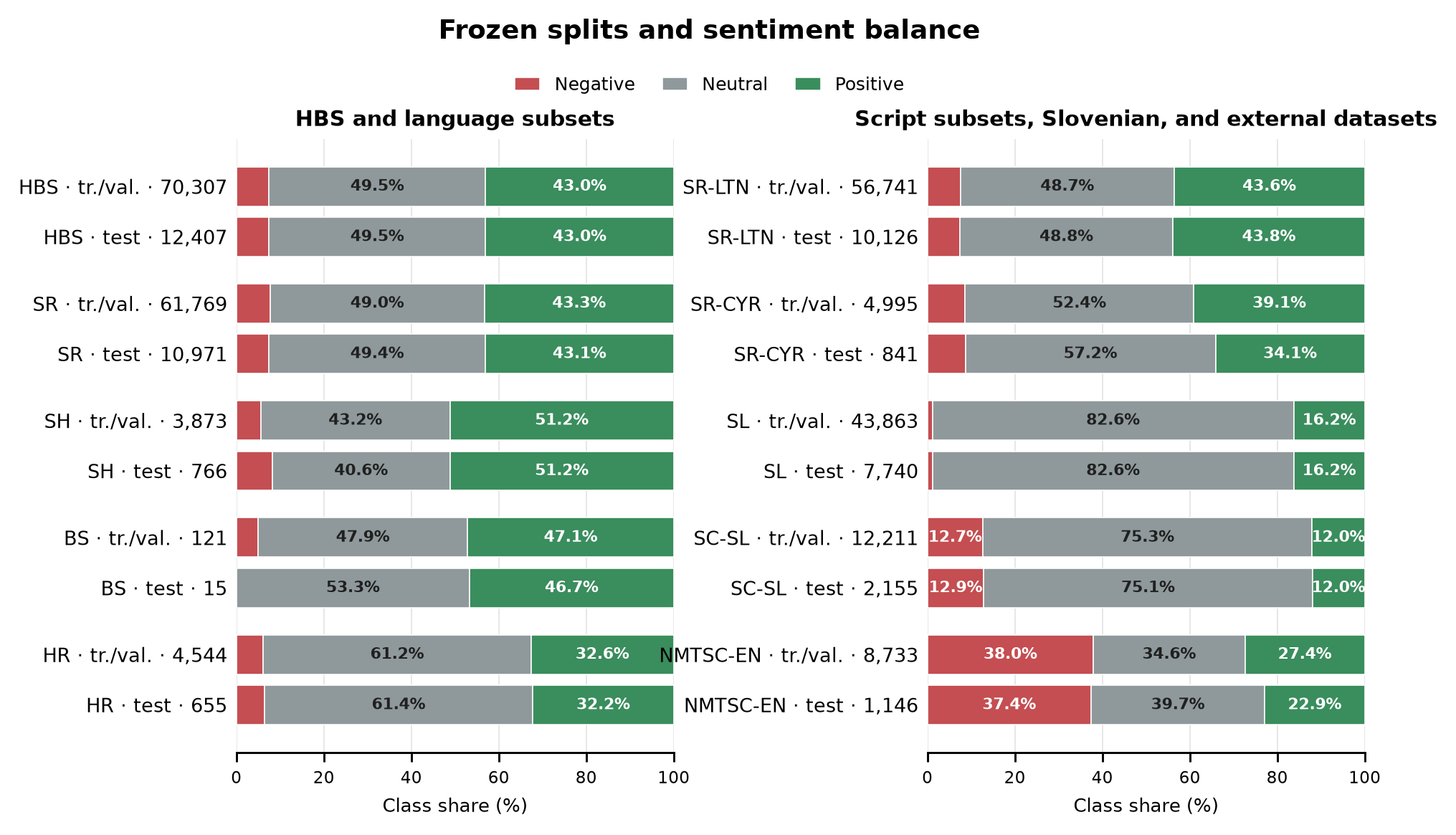}
\caption{Fixed training and test sizes and sentiment balance. AspectBench rows overlap: HBS contains the SR, SH, BS, and HR language subsets, while SR contains the two script subsets. The BS test slice is diagnostic only ($n=15$; no negative example). The Slovenian split also has very few negative examples, whereas NewsMTSC is less skewed. These visible differences limit how directly early learning curves and small-subset scores can be compared.}
\label{fig:full-distributions}
\end{figure*}

The acquisition curves are easier to understand once their starting conditions are visible. Figure~\ref{fig:full-distributions} shows the fixed training and test distributions, including the overlapping AspectBench language and script subsets. SL's early instability is unsurprising once its extremely rare negative class is visible. The same figure also prevents overinterpretation of the Bosnian subset: its test set contains only fifteen records and no negative example, so it can illustrate those records but cannot support a general subgroup claim.

\begin{figure*}[!b]
\centering
\includegraphics[width=.98\textwidth]{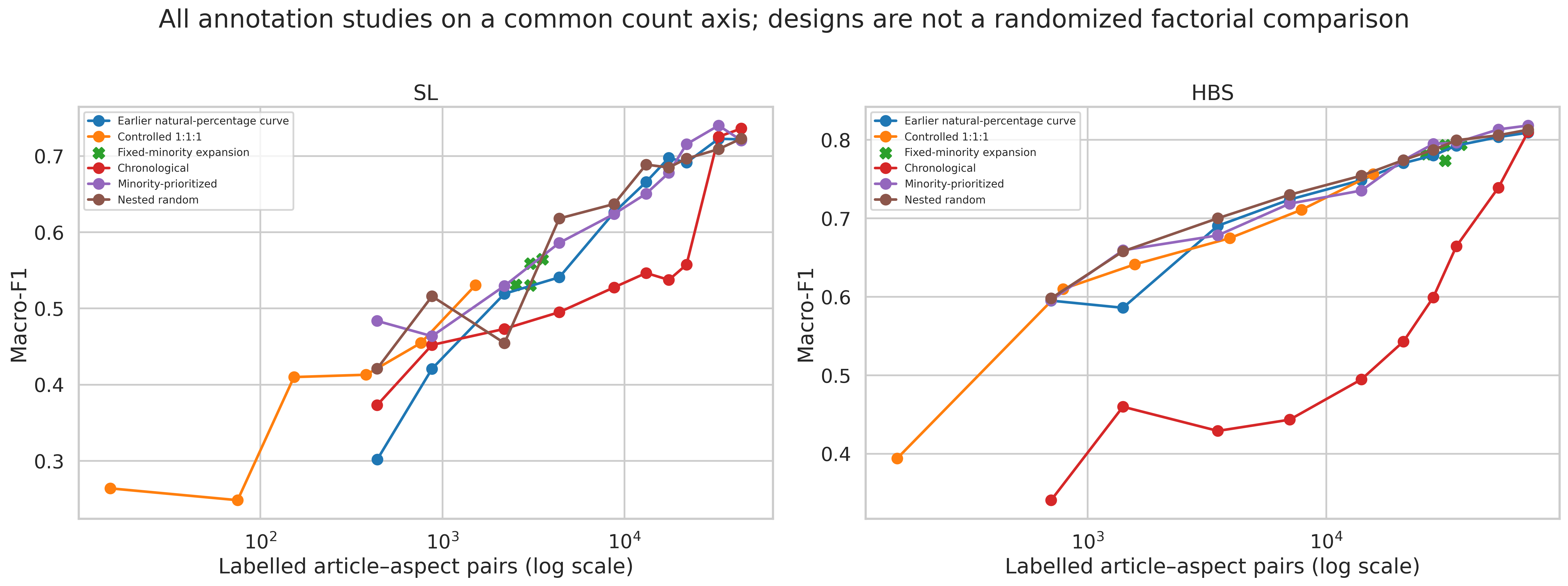}
\caption{All annotation designs plotted against the number of labelled records actually used. Early curves differ, especially for chronological HBS acquisition, but the designs broadly converge at larger budgets. Because several design choices change together, the lines are not a direct controlled comparison of one isolated factor.}
\label{fig:all-acquisition}
\end{figure*}

\begin{figure*}[!b]
\centering
\includegraphics[width=.97\textwidth]{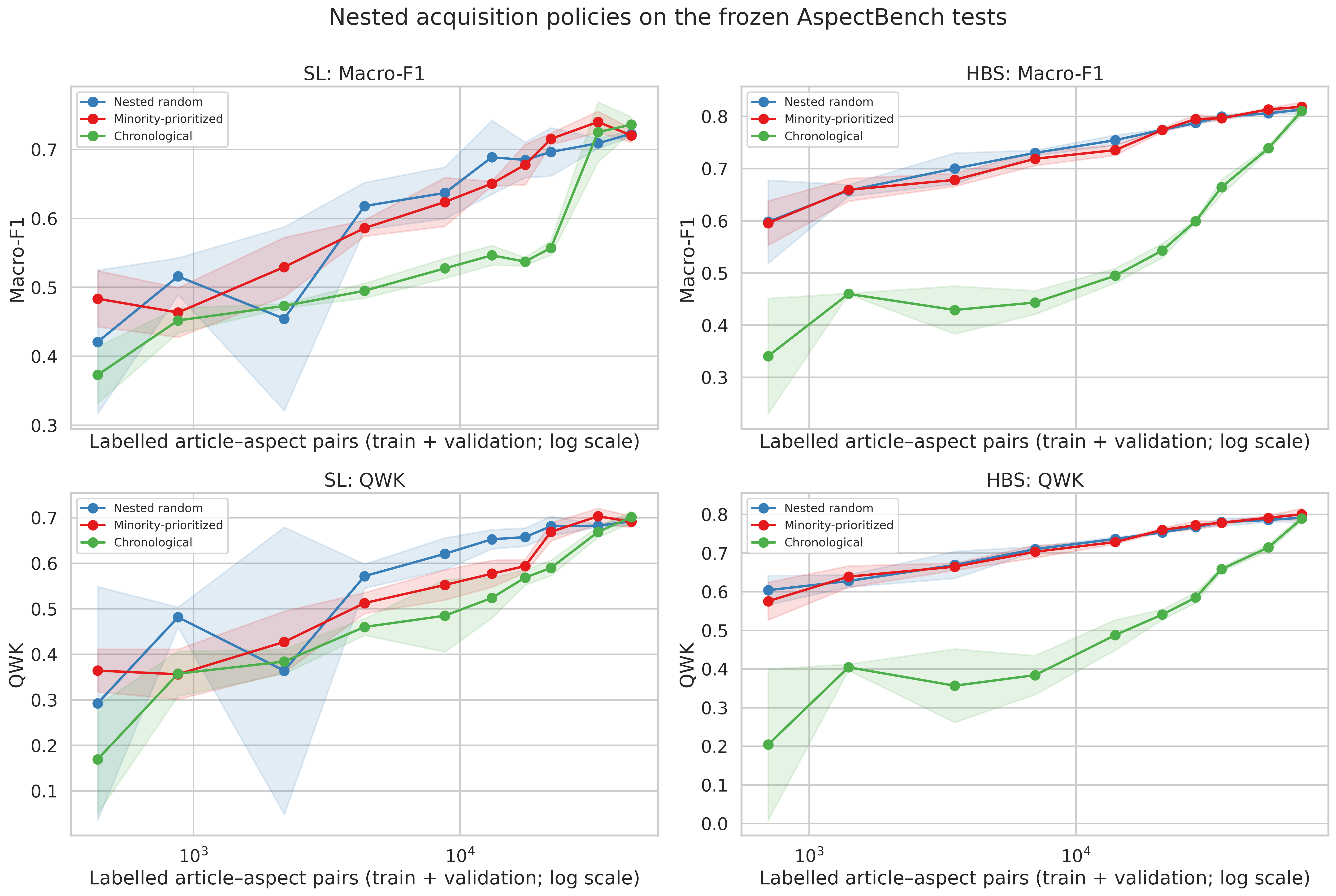}
\caption{Progressively growing acquisition histories with both Macro-F1 and QWK. Chronological HBS acquisition improves more slowly than the random histories, while the broad bands at the smallest SL budgets expose how uncertain a stopping decision is when there are very few negative examples.}
\label{fig:acquisition-full}
\end{figure*}

Figure~\ref{fig:all-acquisition} places the progressively growing experiment beside three earlier designs. Natural-percentage subsets preserve approximately the observed class proportions. Balanced-growth subsets begin with roughly equal numbers from each class. Fixed-minority subsets retain the same rare-class records while adding more neutral and positive examples. These designs help us ask whether a small pilot has enough examples from each class and whether adding common-class records can still improve rare-class performance. They are not prefixes of one acquisition campaign, however, so their lines should not be read as competing sequential policies. Figure~\ref{fig:acquisition-full} also reports QWK and shows why the smallest SL budgets are too unstable for a confident stopping decision.

\begin{figure*}[!t]
\centering
\includegraphics[width=.9\textwidth]{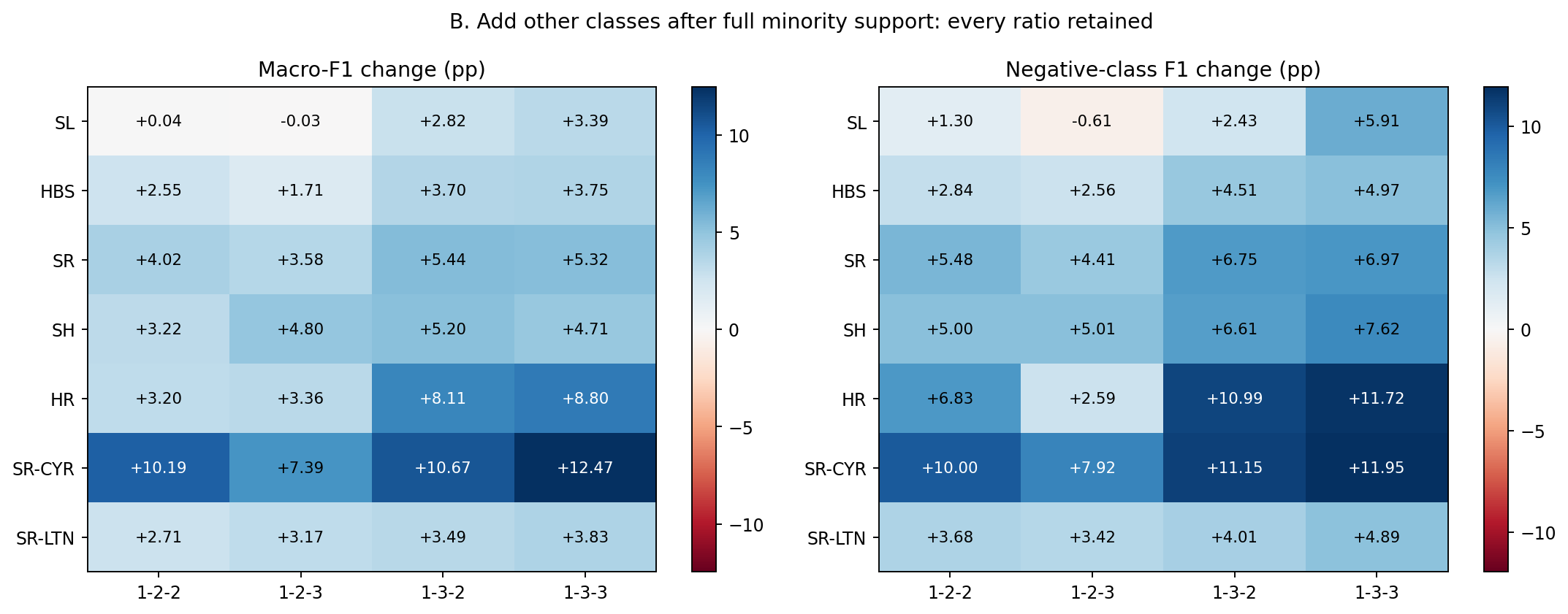}
\caption{Class-level changes after adding common-class records to a fixed negative-class base. Negative-class F1 can improve even though the number of negative training examples is held fixed. The added neutral and positive records also enlarge contextual and entity coverage, so the figure does not isolate class count as the cause.}
\label{fig:class-extension}
\end{figure*}

F1 for the rare negative class can improve after neutral and positive records are added while the negative examples remain fixed (Figure~\ref{fig:class-extension}), because those records also expand target and contextual coverage. We therefore interpret early class balancing as a way to ensure that every class has enough initial examples, not as evidence that a campaign should preserve an equal class ratio or that class composition alone explains the later curve.

\subsection{RQ2 Support: Sensitivity to the Number of Stochastic Predictions}
\label{sec:reduced-mc}

This sensitivity analysis asks how much of the 30-prediction behaviour remains when uncertainty estimation uses fewer forward predictions. SentiCoref and NewsMTSC retain the required pass-level outputs, which we re-evaluate directly. For AspectBench, we use the original masked BERTi\'{c} HBS and SloBERTa SL checkpoints to record ten dropout predictions and one ordinary dropout-disabled prediction from each checkpoint; no model is retrained. A fixed set of pass positions is shared by every record and checkpoint, never chosen separately for individual test items. The reference uses three checkpoints with ten predictions each, while the candidate uses three with three each. Reducing this count changes the averaged label and risk ordering, not the fixed 10\% processing budget. Thus the external encoder results can recompute the complete E1--E2 pipeline, whereas the AspectBench downstream checks change the selected records while retaining stored E2 outputs; stored Gemma answers are not presented as fresh reduced-pass evidence.

\begin{figure*}[!tbp]
\centering
\includegraphics[width=.99\textwidth]{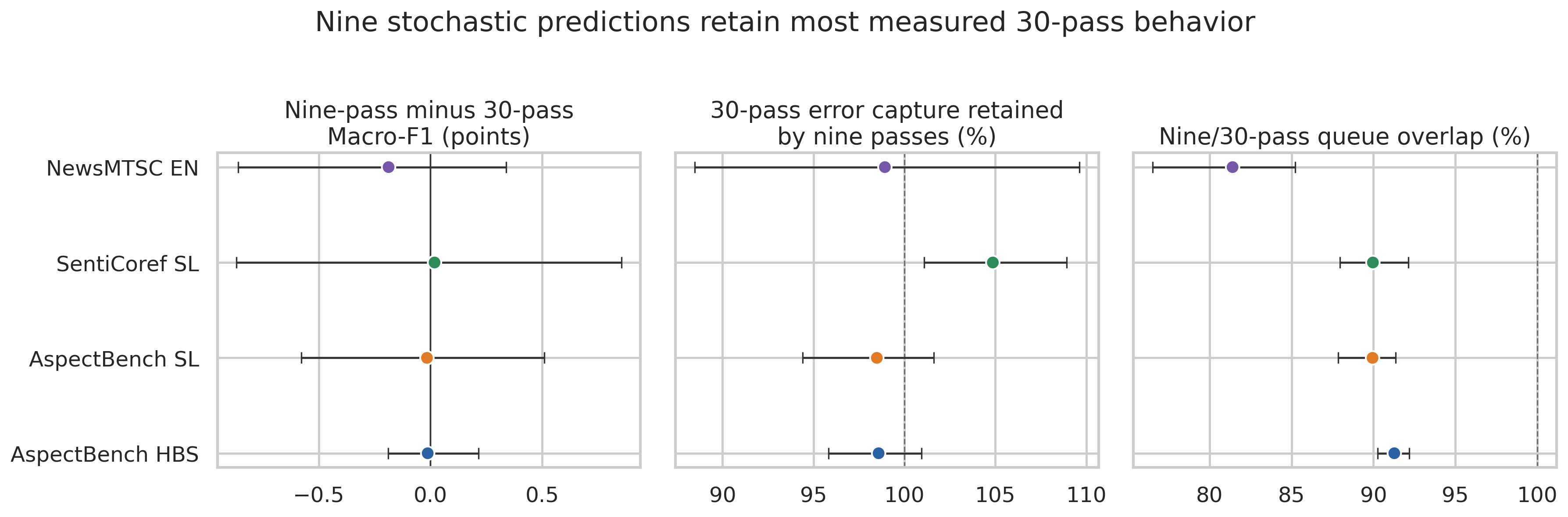}
\caption{Nine stochastic predictions compared with the internally matched 30-prediction reference at a fixed 10\% low-confidence gate. Macro-F1 changes are percentage points. Error-capture retention divides the share of frozen 30-prediction E1 errors selected by the nine-prediction queue by the share selected by the reference queue; it can exceed 100\%. Queue overlap is the intersection divided by the 10\% queue size. AspectBench points average all 120 fixed three-of-ten pass-position subsets and their bars show the 2.5th--97.5th percentile range; external points use the retained pass-level reconstruction and the corresponding subset range. These are sensitivity intervals, not equivalence or non-inferiority tests.}
\label{fig:reduced-mc}
\end{figure*}

Figure~\ref{fig:reduced-mc} shows the four-dataset result. Nine-prediction E1 differs from its 30-prediction reference by -0.012, -0.016, +0.018, and -0.188 Macro-F1 percentage points on AspectBench HBS, AspectBench SL, SentiCoref, and NewsMTSC. It retains 98.6\%, 98.5\%, 104.9\%, and 98.9\% of the reference's captured errors, and its queue overlaps by 91.3\%, 89.9\%, 90.0\%, and 81.4\%. SentiCoref exceeds 100\% error-capture retention because its reduced queue contains more members of the same frozen 30-prediction error set than the reference queue; this does not change the definition of an error or establish a general improvement. AspectBench SL has the widest pass-subset spread: its E1 difference ranges from -0.579 to +0.510 points at the 2.5th and 97.5th percentiles even though the mean is close to zero. No acceptable-loss margin was declared before these test reconstructions, so small means and intervals containing zero do not establish equivalence or non-inferiority.

Repeatedly sampling one trained model does not replace diversity across trained checkpoints. On HBS, ten stochastic predictions from one checkpoint overlap only 56.7\% of the 30-prediction queue, compared with 91.3\% for three predictions from each of three checkpoints; the corresponding SL values are 66.5\% and 89.9\%. A deterministic three-checkpoint probability average is a credible cheaper baseline. Its HBS and SL queue overlaps are 80.5\% and 80.6\%, and it concentrates errors comparably in SL, although the nine-prediction configuration follows the reference queue more closely. Low confidence and predictive entropy behave almost identically on AspectBench, mutual information is less stable, and hard-vote variation ratio is weaker (error-detection AUROC approximately 0.663 on both collections). At a 20\% gate, nine and 30 predictions capture 52.49\% and 52.89\% of fresh-reference HBS errors and 63.87\% and 63.59\% of SL errors, respectively; 10\% remains the primary comparison rate.

Primary-model quality alone does not tell us whether the reduced configuration preserves the downstream action. For SentiCoref and NewsMTSC, matched reduced E1 and E2 probabilities let us recompute the complete encoder pipeline. The nine-prediction pipeline differs from the 30-prediction pipeline by +0.35 points on SentiCoref (95\% paired interval -0.77 to +1.58) and -0.16 on NewsMTSC (-1.12 to +0.74). Relative to its own nine-prediction E1 baseline, averaging changes Macro-F1 by -2.15 points on SentiCoref (-4.44 to -0.20) and +1.55 on NewsMTSC (+0.25 to +2.98). Pipeline differences, action effects, E1-only differences, error capture, and overlap among selected predictions are therefore separate outcomes.

For AspectBench, the stored E2 outputs allow us to test only whether changing the queue changes the result: the E2 decision or historical E1+E2 average remains fixed. At 10\%, the nine-prediction HBS queue yields +0.33 points for E2 replacement and +0.66 for the fixed average; SL yields +1.09 and +1.16. The fixed average changes by only +0.001 points on HBS (95\% interval -0.163 to +0.166) and -0.001 on SL (-0.078 to +0.059) when the nine-prediction queue replaces the 30-prediction queue. This shows that the queue is stable; it is not a fresh end-to-end nine-prediction E1+E2 evaluation. We likewise do not present stored Gemma answers as new nine-prediction evidence because their prompts contain probabilities and uncertainty values from the historical E1 run.

Fresh dropout predictions do not exactly reproduce an older stochastic run. Relative to the historical AspectBench result, the newly exported full-30 reference differs by -0.036 Macro-F1 points on HBS and -0.084 on SL; prediction agreement is 99.20\% and 99.70\%, while 10\% queue overlap is 91.54\% and 90.57\%. The nine-versus-thirty comparison therefore uses the newly generated 30-prediction reference from the same run and does not replace the published historical benchmark elsewhere in the paper. These results make $3\times3$ a practical configuration to validate prospectively, not a universal minimum or optimum.

\FloatBarrier

\subsection{RQ2 Support: Calibration, Ranking, and Corrective Utility}
\label{sec:calibration-ranking}

\begin{figure*}[!t]
\centering
\includegraphics[width=.97\textwidth]{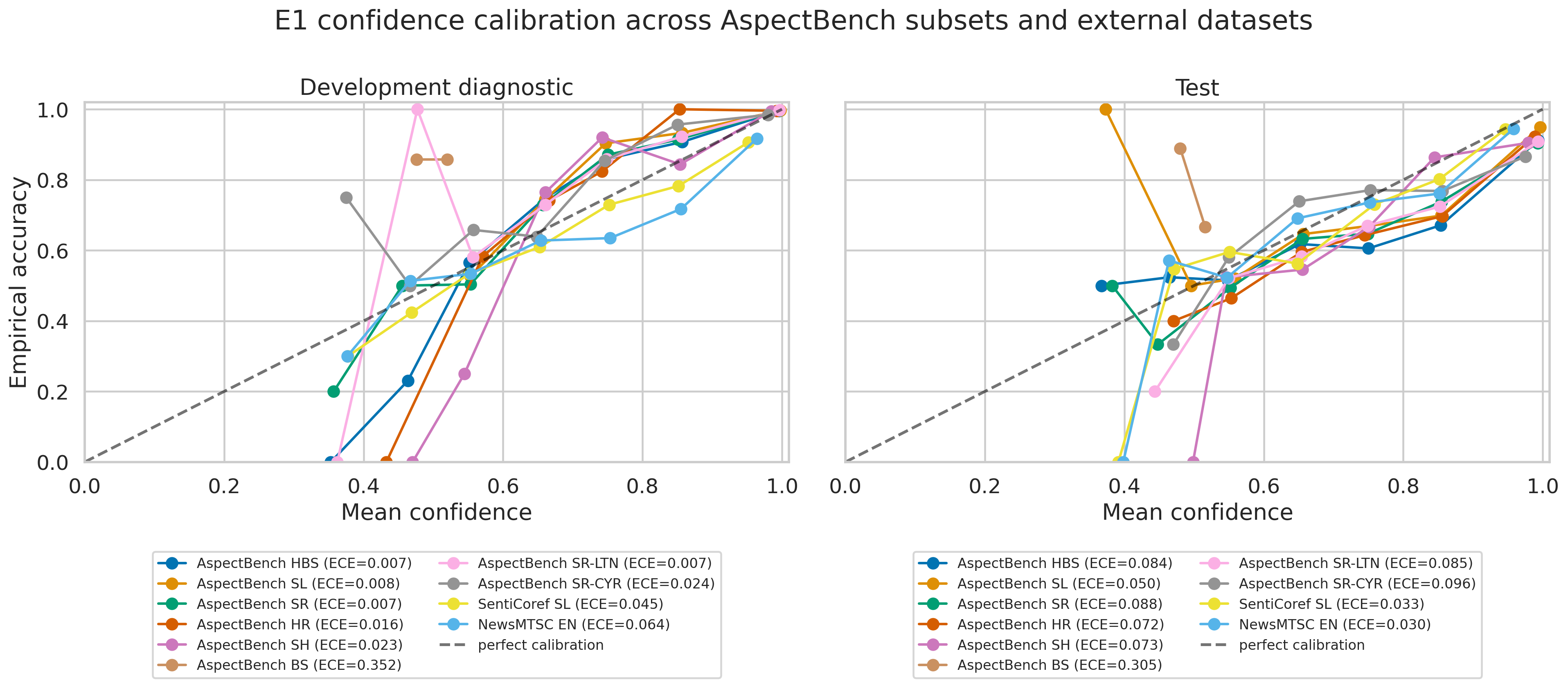}
\caption{How often predictions are correct at different reported confidence levels across AspectBench subsets and the external datasets. Several curves depart visibly from the diagonal, especially in sparse low-confidence regions, so the reported score should not be read as a literal probability that the prediction is correct.}
\label{fig:calibration}
\end{figure*}

For fixed confidence bins $B_b$, top-label expected calibration error is
\begin{equation}
\mathrm{ECE}=\sum_b\frac{|B_b|}{n}
\left|\mathrm{acc}(B_b)-\mathrm{conf}(B_b)\right|.
\end{equation}
ECE asks whether, for example, predictions reported near 0.7 confidence are correct approximately 70\% of the time. Error-detection AUROC asks a different question: after sorting records from low to high confidence, do errors tend to appear before correct predictions? Corrective utility asks a third: once those records have been selected, does a particular action improve the final decisions? A system may perform well on one test and poorly on another.

The external calibration check also depends on the model and whether the target name is hidden. SentiCoref E1 has 3.26 ECE points when masked and 6.02 when unmasked; its E2 changes from 4.84 to 2.98. NewsMTSC E1 changes from 3.30 to 2.97, and E2 from 3.46 to 3.39. Excluding the diagnostic-only Bosnian slice ($n=15$), ECE across the evaluated datasets and subsets ranges from 2.97 to 9.83 points while low-confidence error AUROC ranges from 0.657 to 0.822. The Bosnian slice instead has 36.51 ECE points and 0.528 AUROC, values too unstable to support a subgroup conclusion. We retain these values as a short diagnostic rather than a second reliability figure: Figure~\ref{fig:calibration} already shows the E1 development/test pattern used to interpret the gate. Because the AspectBench subsets overlap, we do not use their correlation to draw a statistical conclusion.

\begin{figure*}[!t]
\centering
\includegraphics[width=.97\textwidth]{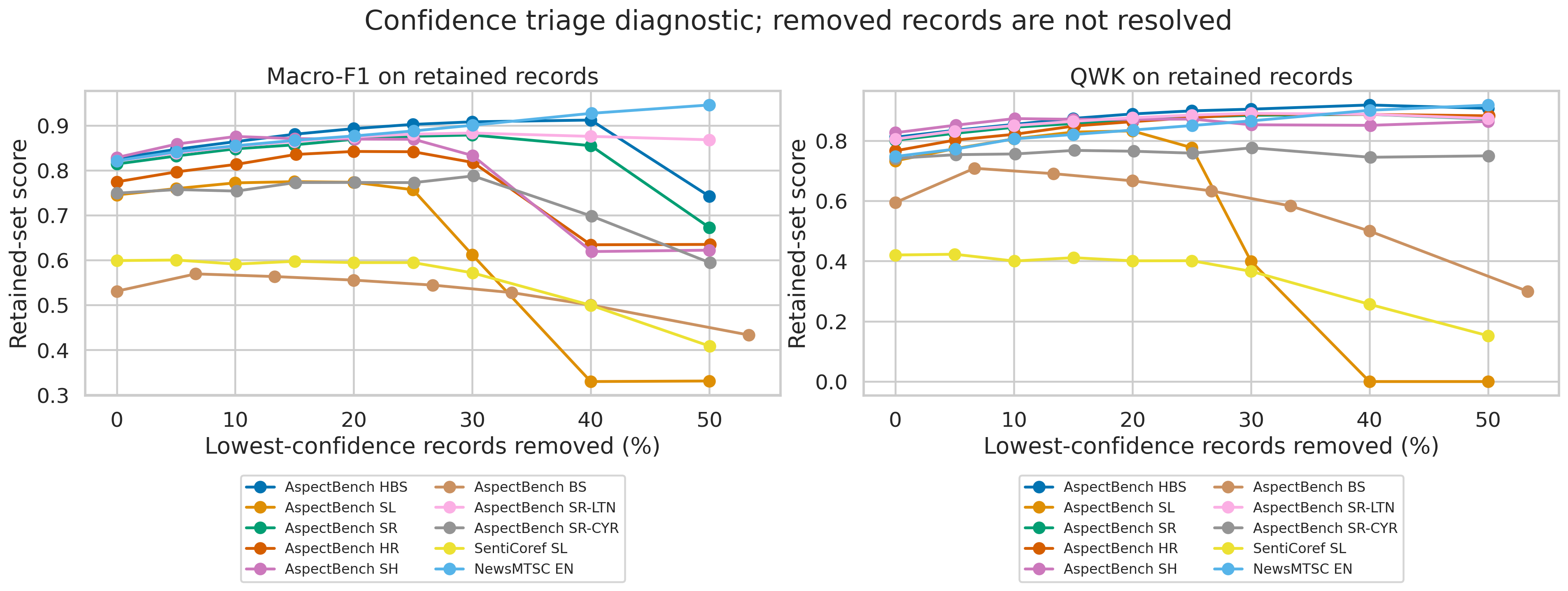}
\caption{Performance on the remaining records as increasingly many low-confidence predictions are withheld. Moderate withholding often improves the retained-set score, but high rates become unstable in small or class-skewed subsets. This figure does not apply a second model to the withheld records and therefore measures triage, not correction.}
\label{fig:triage}
\end{figure*}

\paragraph{Selection rules.}
\phantomsection
\label{sec:detect-selection}

Low confidence provides one reproducible inspection order, but two questions remain. Does a reported confidence of 0.7 mean that roughly 70\% of such predictions are correct? Independently, does the score place errors near the beginning of the ranking? Figures~\ref{fig:calibration} and~\ref{fig:triage} show why these questions should not be conflated: a score can rank errors usefully without being a calibrated estimate of correctness \citep{pmlr-v70-guo17a,geifman2017selective}.

The 80-policy development grids allow the risk signal to change for every action and routing fraction. Their results are mixed rather than evidence of a general action-specific uncertainty signal. Some development-best signal choices differ, but signal preferences and whole-policy rankings are not stable across discovery, confirmation, and test. The experiment therefore supports selecting the signal together with the action on development data; it does not establish that each action has a distinct optimal signal that transfers to unseen data.

The apparent room for perfect correction changes with the ranking rule and queue size. A score derived from Gemma's explanation requires the LLM response before records can be ranked, so it cannot save the call it evaluates. Likewise, retrospectively evaluating subsets of Gemma responses generated for every record does not demonstrate the behaviour of a prospectively deployed selective-calling system. We use these results only to understand how the ranking behaves, not as evidence for a deployable gate.

\subsection{RQ3 Support: Consolidated Strategy Results}
\label{sec:resolve-support}

Table~\ref{tab:full-strategy-appendix} expands the main strategy comparison. Some actions process every record, whereas others reconsider only the preselected 10\%, a distinction that matters for both interpretation and cost. E1 is reported as absolute Macro-F1 and every other value as a point change from E1. Where a cell contains two values, the first hides the target's name and the second retains it. The AspectBench language and script rows locate changes within the aggregate pattern, but their overlapping records are not independent confirmation.

\begin{table}[!tbp]
\centering
\scriptsize
\setlength{\tabcolsep}{2.1pt}
\resizebox{\columnwidth}{!}{%
\begin{tabular}{@{}lrrr rrr@{}}
\toprule
& \multicolumn{3}{c}{Gated E1+E2 average} & \multicolumn{3}{c}{Gated Gemma, target retained} \\
\cmidrule(lr){2-4}\cmidrule(l){5-7}
Dataset & $\Delta$F1 & Correct & Harm & $\Delta$F1 & Correct & Harm \\
\midrule
AspectBench HBS & +0.77 & 198 & 110 & +0.53 & 98 & 67 \\
AspectBench SL & +1.14 & 60 & 36 & +1.23 & 22 & 11 \\
SentiCoref SL & $-2.52$ & 20 & 42 & +0.90 & 11 & 9 \\
NewsMTSC EN & +1.51 & 25 & 12 & +1.28 & 16 & 5 \\
\bottomrule
\end{tabular}
}
\caption{Correction and harm counts for the main same-queue 10\% comparison. Macro-F1 changes are percentage points. Counts must be interpreted together with the sentiment classes changed.}
\label{tab:actions}
\end{table}

\begin{figure*}[!tbp]
\centering
\includegraphics[width=.94\textwidth]{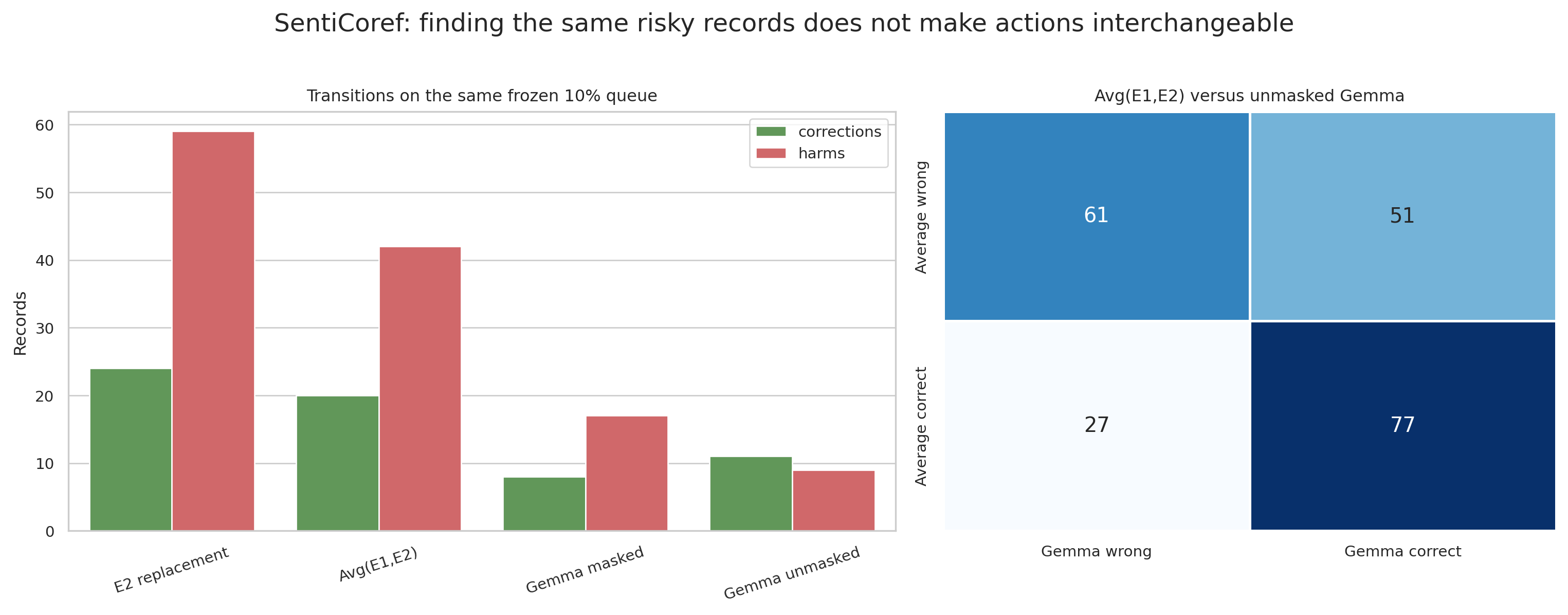}
\caption{Actions applied to the same frozen 216-record SentiCoref queue. Left: corrections of E1 errors and harms to E1-correct predictions. E2 replacement and averaging correct more E1 errors than Gemma but also damage many more correct decisions. Right: record-level correctness of averaging and target-retaining Gemma; Gemma alone is correct on 51 records, whereas averaging alone is correct on 27. These counts describe one frozen queue and do not establish an architectural cause.}
\label{fig:senticoref-reversal}
\end{figure*}

Figure~\ref{fig:senticoref-reversal} explains the SentiCoref reversal at the record level: encoder actions correct more E1 errors, but they also replace many more predictions that E1 had already classified correctly.

Within AspectBench, SR denotes the Serbian subset, SH the dataset's fixed Serbo-Croatian subset, BS the small Bosnian subset, and HR the Croatian subset. SR-CYR and SR-LTN further divide Serbian records written predominantly in Cyrillic and Latin script. These subsets overlap with HBS and with one another.

\begin{table*}[!tbp]
\centering
\scriptsize
\setlength{\tabcolsep}{2.7pt}
\begin{tabular}{@{}lrrrrrrr@{}}
\toprule
Evaluation & E1 & \shortstack{E2 on\\every record} & \shortstack{Gemma without\\E1 information} &
\shortstack{E1+E2 average\\on every record} & \shortstack{Gated E1+Gemma\\at 10\%} &
\shortstack{E1+Gemma on\\every record} & \shortstack{Perfect correction\\at 10\%} \\
\midrule
AspectBench HBS    & 82.63 & +0.33 & $-22.90$          & +1.22 & +0.58 / +0.53 & $-3.82$ / $-1.13$ & +5.75 \\
AspectBench SL     & 74.53 & +0.72 & $-31.98$          & +1.27 & +1.06 / +1.23 & $-1.44$ / $-3.80$ & +10.36 \\
AspectBench SR     & 81.42 & +1.31 & $-22.43$ / $-19.89$ & +2.25 & +0.00 / +0.29 & $-2.07$ / $-1.02$ & +5.90 \\
AspectBench SH     & 82.86 & +3.53 & $-12.01$ / $-14.18$ & +4.10 & +0.99 / +1.55 & $-0.21$ / +0.72 & +6.13 \\
AspectBench HR     & 77.45 & +4.26 & $-26.37$ / $-23.99$ & +3.73 & $-0.28$ / $-0.76$ & $-1.15$ / $-1.74$ & +6.76 \\
AspectBench SR-CYR & 74.93 & +7.10 & $-20.83$ / $-15.21$ & +7.41 & +0.01 / +0.00 & $-2.19$ / +0.52 & +4.29 \\
AspectBench SR-LTN & 82.10 & +0.69 & $-25.09$ / $-20.62$ & +1.64 & +0.14 / +0.61 & $-3.47$ / $-1.11$ & +5.54 \\
SentiCoref SL      & 59.91 & $-6.41$ & $-14.31$ / $-10.70$ & $-2.92$ & +0.09 / +0.90 & $-0.14$ / +0.40 & +9.14 \\
NewsMTSC EN        & 82.14 & +0.48 & $-9.30$ / $-5.46$ & +1.89 & +0.27 / +1.28 & $-0.70$ / +0.50 & +5.05 \\
\bottomrule
\end{tabular}
\caption{Consolidated strategy results. E1 is absolute Macro-F1; all remaining cells are point changes from E1. A slash separates Gemma results when the target's name is hidden or retained. Gated strategies and perfect correction use the least-confident 10\% of E1 predictions, whereas the other strategies process every record. AspectBench rows overlap, and the resource budgets of the strategies differ.}
\label{tab:full-strategy-appendix}
\end{table*}

\begin{figure*}[t]
\centering
\includegraphics[width=.82\textwidth]{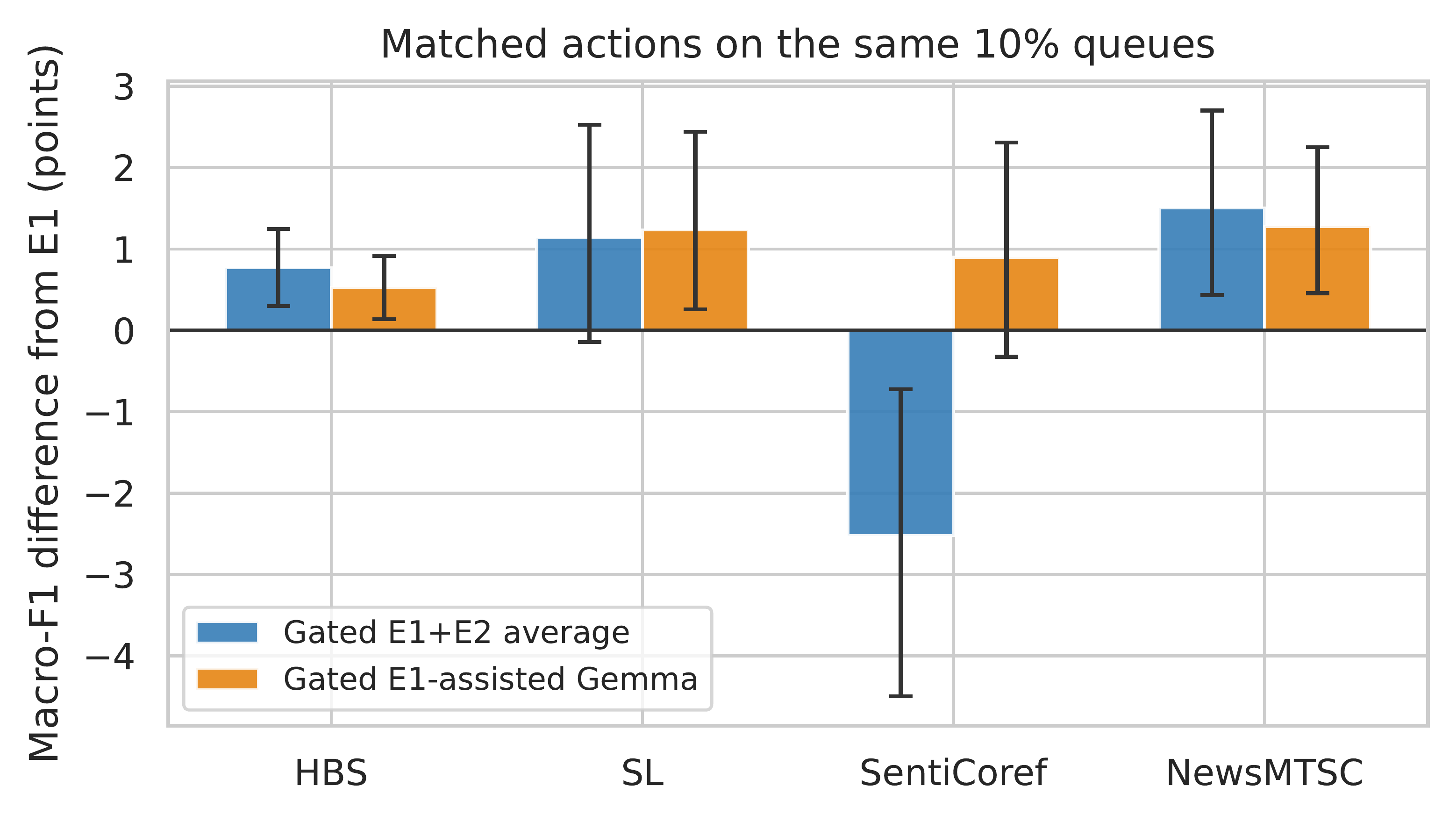}
\caption{Two actions applied to the same least-confident 10\% of document--entity ELSA records. Bars show Macro-F1 change from E1 and intervals come from paired source-document resampling. Encoder averaging improves AspectBench and NewsMTSC but harms SentiCoref. Gemma changes less in magnitude across these four comparisons, although the SentiCoref interval includes both improvement and deterioration. The figure supports RQ3: an error-rich queue does not identify the action that should receive it.}
\label{fig:strategy}
\end{figure*}

Table~\ref{tab:full-strategy-appendix} and Figure~\ref{fig:strategy} show the same broad pattern at different levels of detail. Averaging the two encoders' probabilities on every record captures most of the encoder benefit across AspectBench and remains positive on NewsMTSC, while SentiCoref is a clear counterexample. Selective Gemma assistance can add value on those same predictions, but the result depends on the dataset and whether the target name is retained. The perfect-correction column shows that finding many E1 errors can expose substantially more opportunity than any tested action realises.

\subsection{RQ3 Support: Applying Different Actions to the Same Selected Records}

\begin{figure*}[!tbp]
\centering
\includegraphics[width=.97\textwidth]{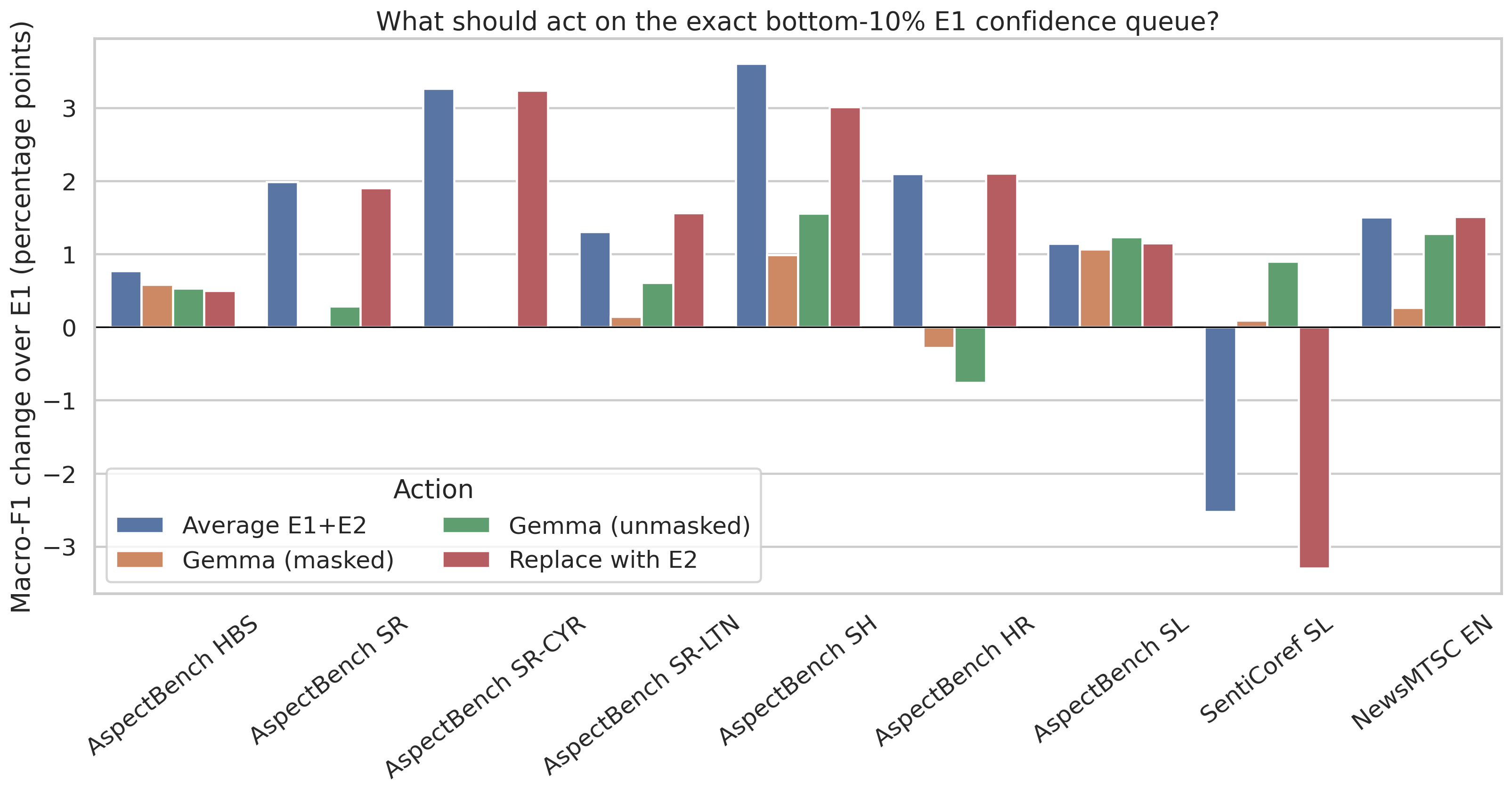}
\caption{Actions on the same least-confident 10\% of E1 predictions. Probability averaging is positive for most AspectBench evaluations and NewsMTSC but negative on SentiCoref; Gemma's changes are smaller and depend on whether the target name is retained. Holding the queue fixed shows that action choice, not only error detection, determines the result.}
\label{fig:same-queue-all}
\end{figure*}

\begin{figure*}[!tbp]
\centering
\includegraphics[width=.9\textwidth]{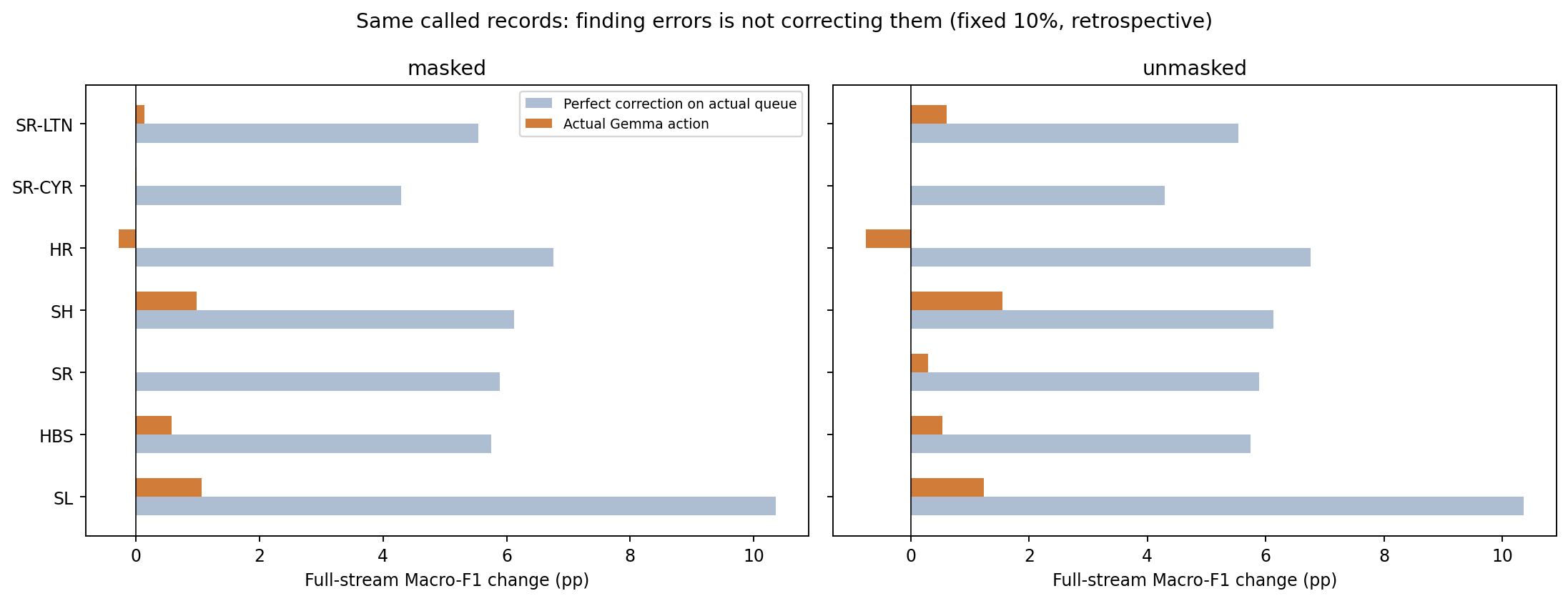}
\caption{Maximum possible improvement and the improvement actually achieved by Gemma on the same selected records. The realised orange bars are much smaller than the blue oracle headroom and are sometimes negative, which makes the gap between finding errors and correcting them visible.}
\label{fig:headroom-action}
\end{figure*}

The main 10\% comparison gives every action the same predictions, so none benefits from receiving an easier subset. Figure~\ref{fig:same-queue-all} extends that comparison across every available AspectBench language or script subset and the two external datasets. Figure~\ref{fig:headroom-action} then places the errors available to correct beside the improvement Gemma actually realises. The gap explains why detection and resolution must be evaluated separately: the selected predictions can include many E1 errors while the proposed action leaves most unresolved or introduces new ones. The HBS subsets locate this pattern within an overlapping collection; they do not independently confirm it.

\begin{figure*}[!tbp]
\centering
\includegraphics[width=.96\textwidth]{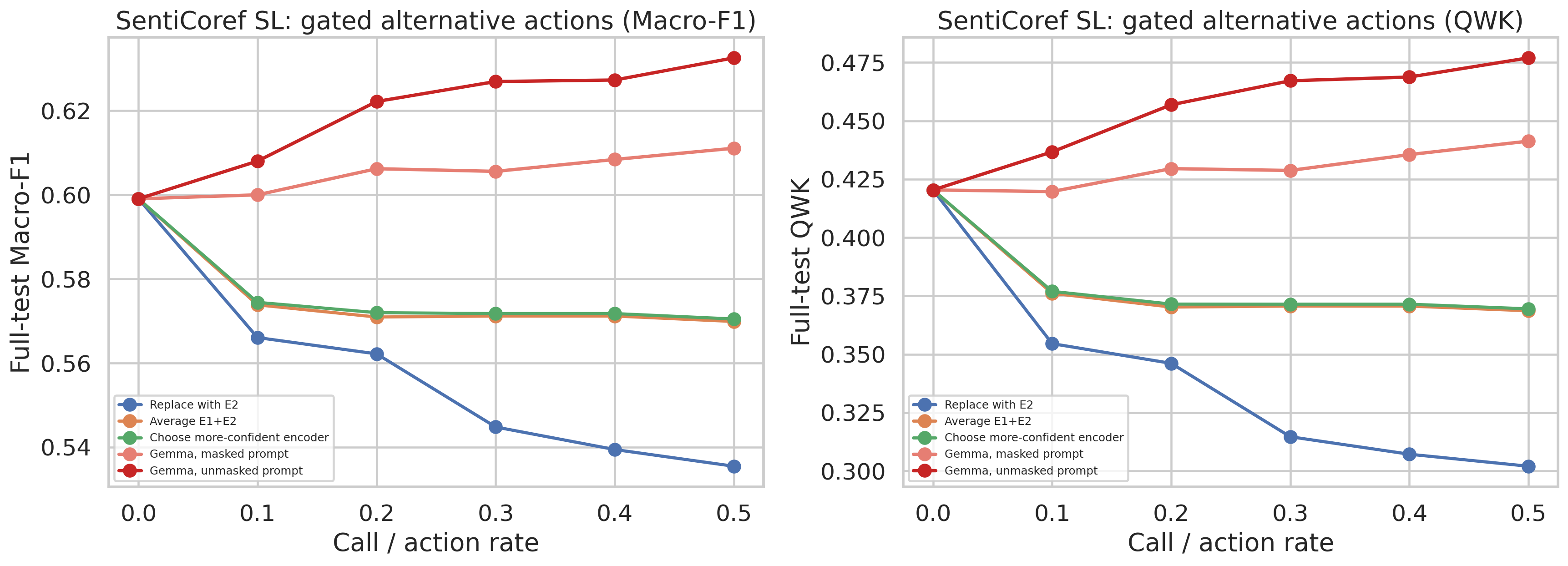}
\caption{SentiCoref actions across progressively larger sets of low-confidence predictions. Replacing E1 with E2 or averaging the encoders becomes increasingly harmful as coverage grows, whereas Gemma with the target name retained provides a modest gain at some intermediate rates.}
\label{fig:senticoref-actions}
\end{figure*}

\begin{figure*}[!tbp]
\centering
\includegraphics[width=.96\textwidth]{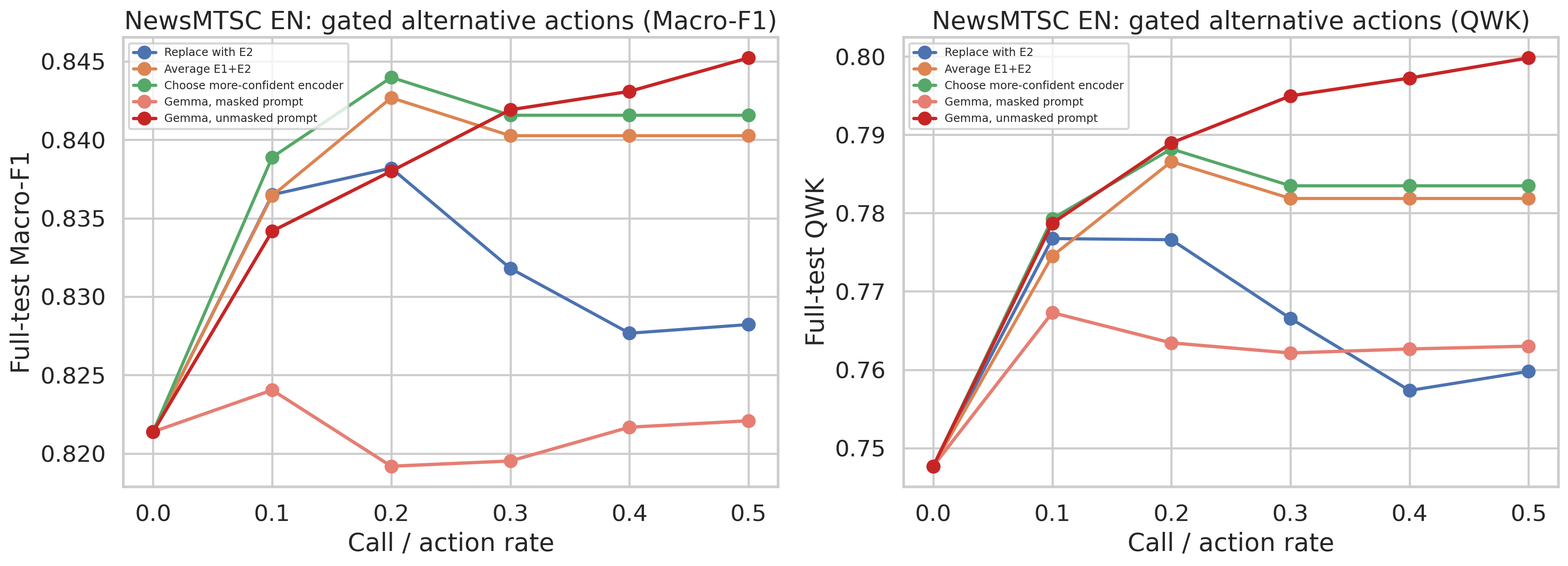}
\caption{NewsMTSC actions across progressively larger sets of low-confidence predictions. Encoder replacement and averaging help most at low or moderate coverage, while the unmasked Gemma curve continues to improve across the plotted range. These test-set comparisons describe sensitivity to queue size; they do not select a deployment threshold.}
\label{fig:newsmtsc-actions}
\end{figure*}

Figures~\ref{fig:senticoref-actions} and~\ref{fig:newsmtsc-actions} then expand the fixed comparison to progressively larger selected sets containing 10 through 50\% of the test set. NewsMTSC shows useful gains across a broader range of actions. SentiCoref does not: E2 remains harmful, while Gemma with the target name retained becomes useful at 20 and 30\% of the records. These curves reveal sensitivity to the fraction sent to another model, but the larger rates were not chosen using an independent development set. We therefore use them to understand the action, not to declare 20 or 30\% a deployment threshold.

\subsection{RQ3 Support: Choosing a Gated-Escalation Rate}
\label{sec:gate-rate}

Ten percent is a controlled comparison budget, not an ideal deployment rate. To examine larger budgets, a later experiment expands each confidence-ranked selection from 10\% to 50\% while reusing stored Gemma answers. The five subsets within a curve are nested, so each step shows what the newly admitted records contribute. AspectBench uses a later Gemma generation than the exact matched-action comparison, whereas the external 10\% answers are reused. We therefore use the exact run to compare actions and the later run to describe the shape of the coverage curve; they are not repeated samples from one stochastic experiment.

\begin{figure*}[!tbp]
\centering
\includegraphics[width=.98\textwidth]{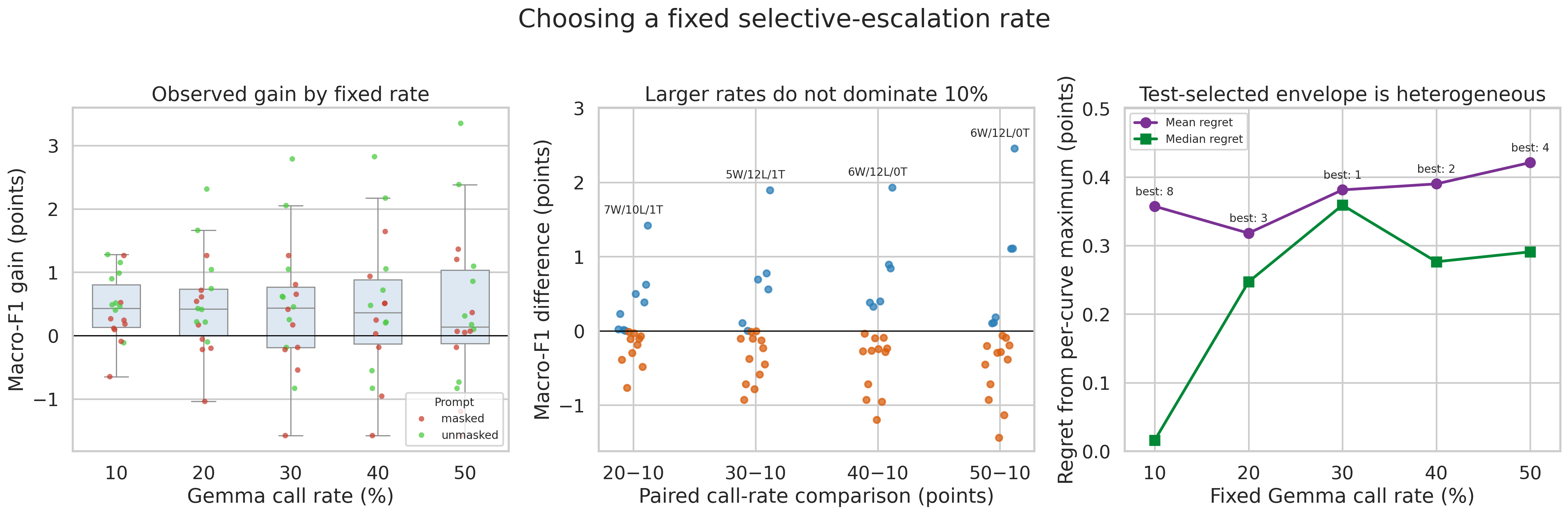}
\caption{Comparison of fixed Gemma call rates in the internally nested 10\% to 50\% experiment. The left panel shows the observed Macro-F1 gains, and the middle panel compares every larger rate with 10\%. The right panel summarises regret from the best observed point on each test curve. These test-selected maxima describe heterogeneity; they are not deployment thresholds.}
\label{fig:fixed-rate-policy}
\end{figure*}

Figure~\ref{fig:fixed-rate-policy} shows why a larger queue is not automatically better. Across 18 dataset and prompt curves, 20\% improves on 10\% in seven comparisons, performs worse in ten, and ties in one. The mean difference between 20\% and 10\% is +0.039 Macro-F1 points, while the median is -0.023 points; neither the paired Wilcoxon test ($p=.782$) nor the two-sided sign test ($p=.629$) supports a general 20\% advantage. These tests summarise the observed curves descriptively rather than providing population-level confirmatory evidence: the curves are dependent because AspectBench language and script subsets overlap and prompt variants share records and stored model outputs. Ten percent is the best or tied rate on eight curves, compared with three at 20\%, one at 30\%, two at 40\%, and four at 50\%. Twenty percent has slightly lower mean regret from each curve's test-selected maximum, largely because a few SentiCoref and NewsMTSC curves benefit from broader escalation. Ten percent has much lower median regret, is most frequently optimal, and uses half as many Gemma calls.

The successive queue bands clarify the cost of expanding coverage. The first band, from 0\% to 10\%, averages +0.445 Macro-F1 points and 6.22 more corrections than harms. The band from 10\% to 20\% adds only +0.039 points and averages 1.72 more harms than corrections. The mean incremental changes for the bands from 20\% to 30\%, 30\% to 40\%, and 40\% to 50\% are -0.063, -0.009, and -0.031 Macro-F1 points, with negative mean net corrections in all three. Macro-F1 can nevertheless improve when a small number of minority-class corrections outweigh more numerous majority-class changes, so class-level scores and counts of corrections and harms must be read together.

\begin{figure*}[!tbp]
\centering
\includegraphics[width=.96\textwidth]{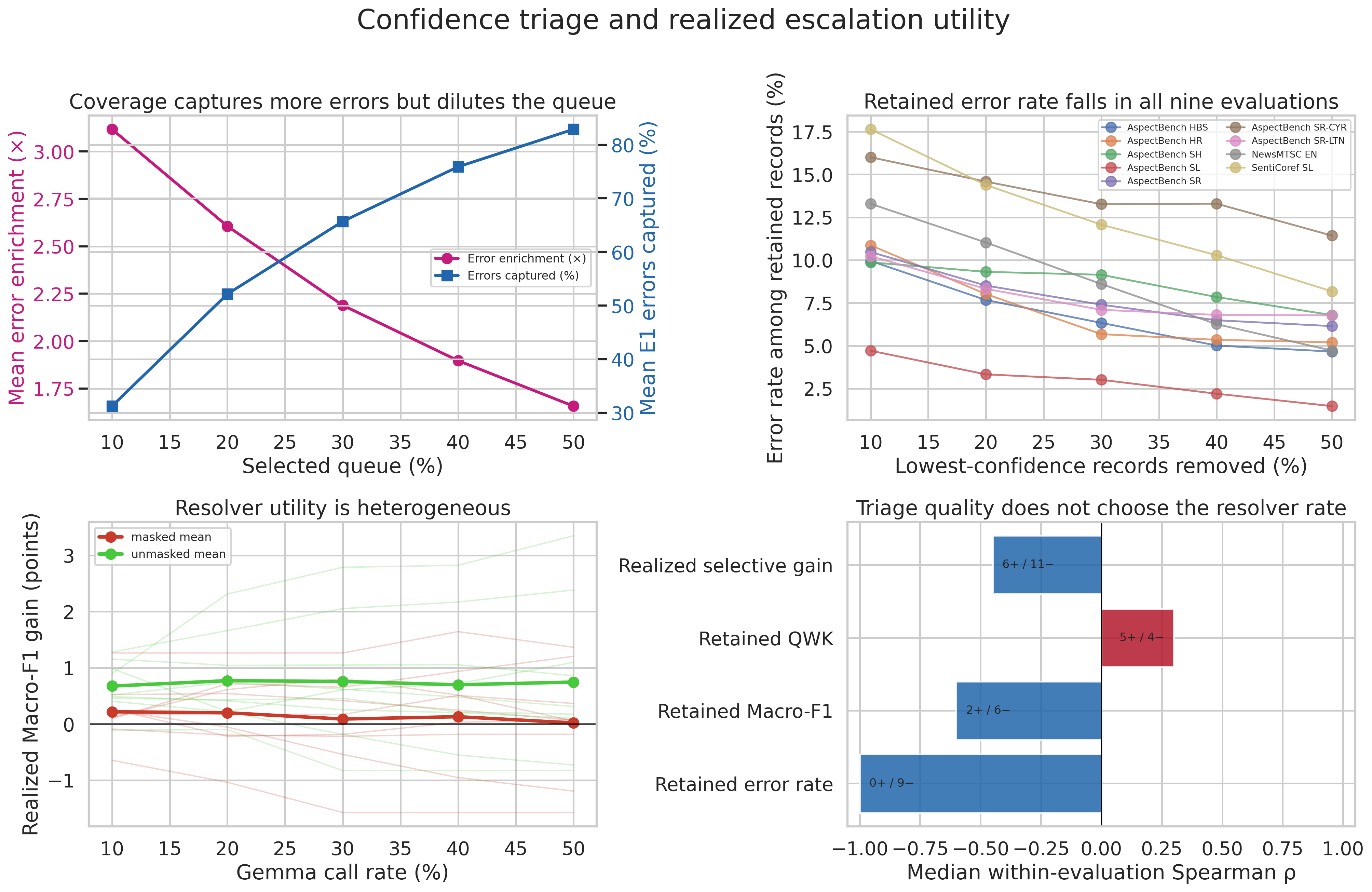}
\caption{Confidence triage and realised gated-escalation utility. As coverage grows, the queue captures more E1 errors but becomes less concentrated for errors. Removing low-confidence records reduces the retained error rate in all nine evaluated collections and subsets, yet neither the retained-set curves nor the gate rate determine how much Macro-F1 Gemma ultimately recovers.}
\label{fig:confidence-triage-utility}
\end{figure*}

Confidence continues to rank difficult records usefully as the queue grows, but the queue becomes less concentrated for errors. Figure~\ref{fig:confidence-triage-utility} shows mean error enrichment falling from 3.12 times at 10\% to 1.66 times at 50\%, while the share of all E1 errors captured rises from 31.25\% to 82.89\%. The error rate among records left with E1 falls in all nine evaluated collections and subsets: the median within-evaluation Spearman correlation with removal rate is $-1.0$, and all nine directions are negative (two-sided sign test $p=.0039$).

That monotonic triage pattern does not determine how far the Gemma queue should extend. Across the 18 dataset and prompt curves, the median correlation between call rate and realised Macro-F1 gain is $-.45$: eleven correlations are negative, six positive, and one zero (sign test $p=.332$). The median correlation between retained-set Macro-F1 and realised gain is $.05$. The retained-set curve evaluates the records left with E1, whereas the resolver acts on the records removed from E1. A useful gate can therefore expose errors without identifying a universally useful resolver rate.

For a new domain, these results motivate comparing 10\% and 20\% first on held-out validation while accounting explicitly for call cost. Larger fractions should be considered only while validation utility continues to rise, and the chosen fraction should be frozen before final test evaluation. This is a procedure suggested by the observed heterogeneity, not evidence that 10\% or 20\% is universally preferable.

\subsection{RQ3 Support: Combining Encoders and Language Data}
\label{sec:encoder-language}

\begin{figure*}[!t]
\centering
\includegraphics[width=.88\textwidth]{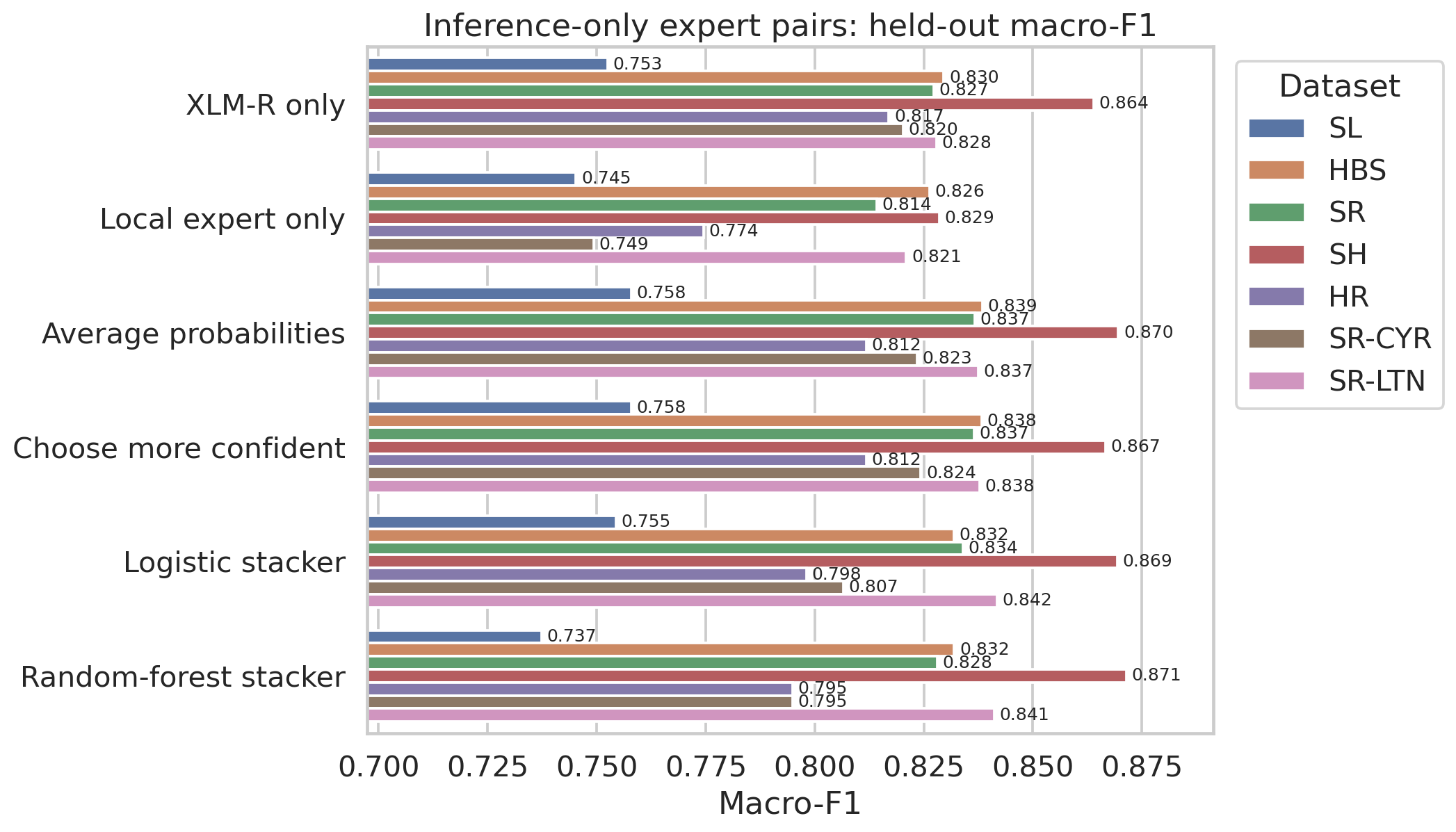}
\caption{E1, E2, and encoder combinations across overlapping AspectBench language and script subsets. Probability averaging and learned stackers are often strong, but the leading combination changes across subsets; no single combiner dominates every subset.}
\label{fig:encoder-pair}
\end{figure*}

A probability average requires no separately trained combiner. On AspectBench and NewsMTSC it captures most of the observed benefit from combining the evaluated encoders. Figure~\ref{fig:encoder-pair} widens this comparison across model pairs. The more elaborate combinations do not improve clearly or consistently enough here to justify their additional training and inference by default.

\begin{figure*}[!t]
\centering
\includegraphics[width=.70\textwidth]{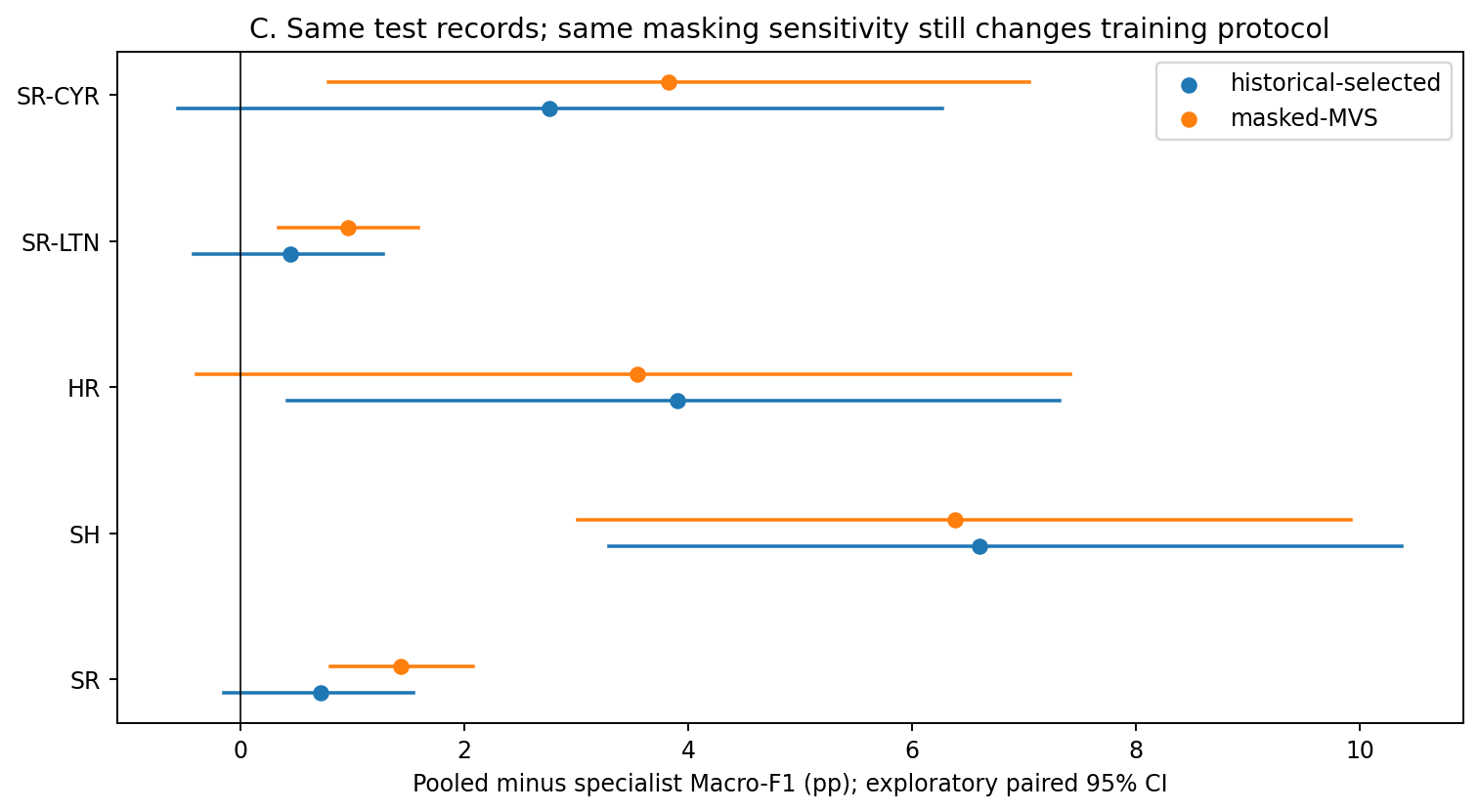}
\caption{The pooled HBS model compared with language-specific models on the same test records. The pooled model shows larger gains for SH, HR, and the Serbian Cyrillic subset than for several other subsets, but heterogeneous intervals and different training pools prevent a causal claim that pooling itself produces the gains.}
\label{fig:pooling}
\end{figure*}

Figure~\ref{fig:pooling} asks a different question: it compares a model trained on the combined HBS languages with models trained for individual languages, using the same test records. These experiments do not hold training-set size, language coverage, and training procedure constant one at a time, so they do not prove that combining the HBS training data is intrinsically superior.

More precisely, the pooling analysis evaluates one BERTi\'{c} model trained on the complete HBS pool against separately trained specialists after restricting both predictions to the same SR, SH, HR, SR-LTN, or SR-CYR test records. It reports historical selected specialists and existing masked minimum-viable-set specialists, but it does not match their training volumes or protocols. We did not train HR+SR, HR+BS, HR+SH, or any other bespoke subset combination; the presence of those records inside HBS is not evidence that such combinations were tested. BS is absent from this comparison because its 15-record test slice has no negative example and does not support a stable three-class comparison.

\subsection{RQ3 Support: Reasoning and Structured-Output Analyses}
\label{sec:reasoning-output}

\begin{figure*}[!b]
\centering
\includegraphics[width=.82\textwidth]{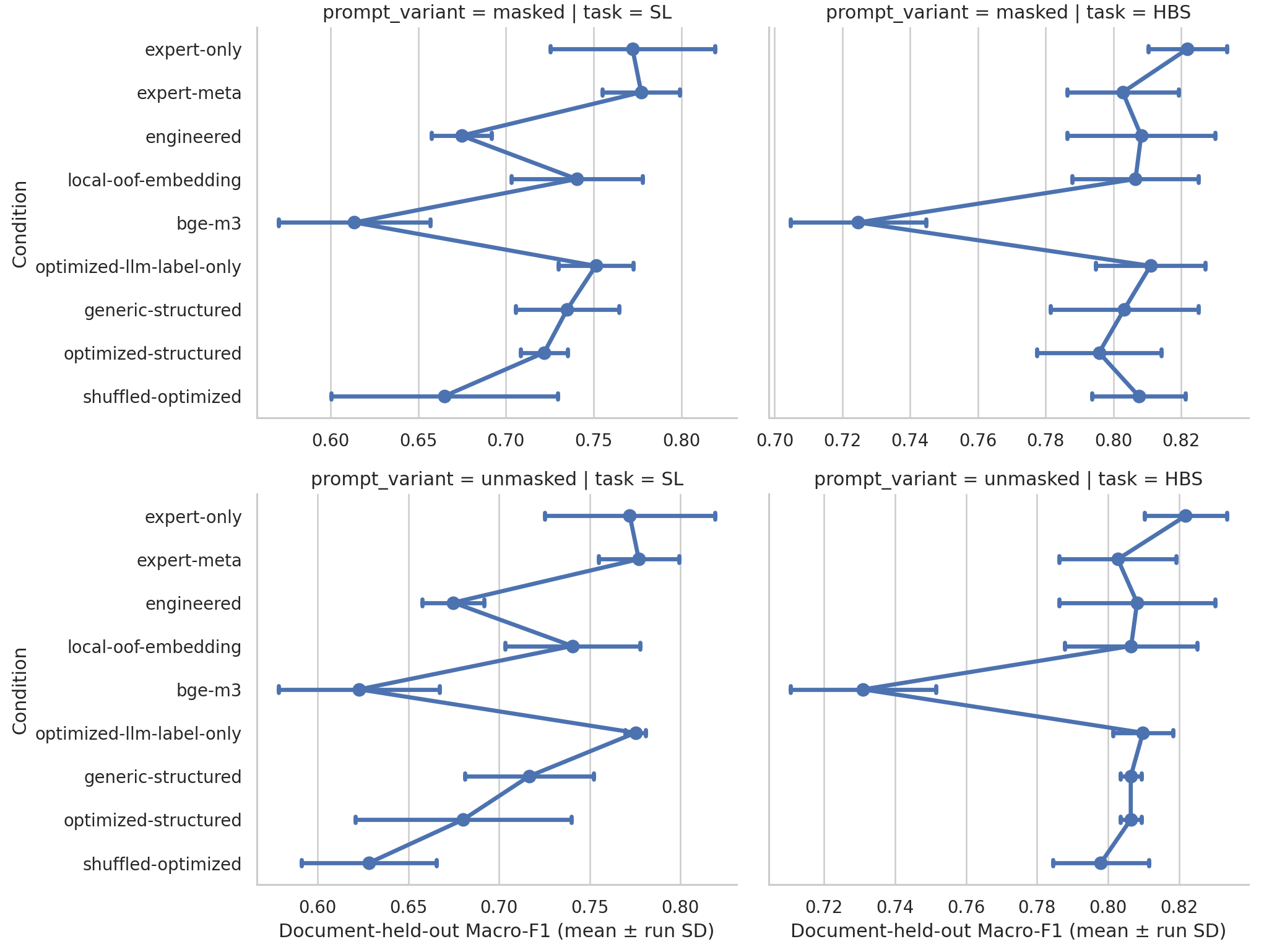}
\caption{Comparison of variants that use separate fields from the LLM output. The experiment does not show a stable gain over the simpler procedure.}
\label{fig:structured}
\end{figure*}

The LLM programs return an explanation and structured fields alongside the sentiment label. Because an articulate explanation is not itself evidence that the label is correct, we test whether those fields help verify the answer. Figure~\ref{fig:structured} compares variants that use separate response fields as model inputs; the variants are fixed before test evaluation.

The feature pilot provides no reliable gain. Its inputs are also available only after Gemma has been called, so they cannot remove the generation cost. This result does not show that explanations or structured outputs can never help. It shows that these particular features should not be treated as a validation mechanism without stronger results on separate development data. This cautious interpretation is consistent with evidence that self-explanations are not always faithful \citep{madsen-etal-2024-self} and that intermediate reasoning is not always used to produce the final answer \citep{paul-etal-2024-making}.

\subsection{RQ4 Select Protocol: Development Selection, Contextual Routing, and Robustness}
\label{sec:policy-transfer-protocol}

The complete policy-selection experiment is run on SentiCoref and NewsMTSC. Its purpose is to choose the risk signal, processing fraction, and corrective action from development records before measuring the retained choice on test. Discovery ranks the candidates, confirmation can only veto the discovery winner, and test evaluates the retained policy. An earlier development choice of encoder-input variant means that confirmation is not independent of every upstream modelling decision. We first remove every source document used during DSPy optimisation, preventing prompt calibration and policy development from sharing records. E1 and E2 development predictions follow the same source-document-group out-of-fold construction: a record is predicted only by checkpoints for which its complete source document belonged to the validation fold and was absent from training. The audit found no E1/E2 identifier, group, or gold-label mismatch and no overlap among source documents used for DSPy optimisation, discovery, confirmation, or test. In-sample development predictions would make both prediction selection and corrective action appear easier to choose than they are.

Whole source documents, rather than individual article--entity pairs, are assigned deterministically to a 70\% discovery partition and a 30\% confirmation partition. Most source groups are thereby available for the 80-policy search, while disjoint records remain for a final point-estimate veto. This is a pragmatic design choice, not a split selected by a power calculation. Target-masked versus target-retaining encoder-input variants were selected by held-out development Macro-F1 before this division. Confirmation is therefore independent of the policy search, but conditional on that earlier modelling choice. Test records and labels enter neither checkpoint training, variant selection, nor policy selection. Final test predictions use the three-checkpoint ensembles; ``out of fold'' describes only the development predictions used for selection.

The grid contains $4\times4\times5=80$ policies. In each one, the signal ranks the records, a fixed-fraction (top-$k$) rule selects exactly the requested share, and the action replaces E1 only for those records. Discovery proposes the policy with the largest gain over E1. Confirmation retains that exact policy only if its point-estimate gain remains positive; otherwise the final rule makes no additional call and keeps E1. No minimum effect or uncertainty threshold was predeclared, so this is a safety-oriented veto rather than formal statistical confirmation. Ties prefer a lower call rate and then, in order, E2 replacement, the more-confident encoder, averaging, masked Gemma, and target-retaining Gemma.

Top-$k$ selection requires a batch or time window because the requested share is defined relative to other records. For comparison, $\pi_{\tau}=(s,\tau,a)$ stores the discovery-set risk-score threshold associated with that share. This numerical-threshold rule can process one record at a time, but the fraction sent to the second model may change when the score distribution shifts.

The dataset-level policy above gives every selected prediction the same action. The contextual comparison asks a different question: can development data choose an action separately for each record? We fix the candidate region to the lowest-confidence 30\% and retain E1 elsewhere. Within that region, five independent class-balanced logistic regressions estimate the correctness of retaining E1, replacing it with E2, averaging, using masked Gemma, or using unmasked Gemma. The E1 feature regime contains E1 probabilities, confidence, margin, entropy, mutual information, variation ratio, and predicted class. The E1+E2 regime adds E2 probabilities, confidence, disagreement, and absolute probability differences. The router takes the action with the greatest estimated probability of correctness. It does not directly optimise non-decomposable Macro-F1, and several actions can be correct on the same record, so there is no unique gold action label.

The robustness analysis keeps every policy choice fixed. We apply the fixed-fraction and numerical-threshold versions to 30 predetermined resamples. One analysis retains 25\%, 50\%, or 75\% of source groups without replacement. Another samples source groups with replacement to create approximately observed, balanced, neutral-heavy, polar-heavy, or negative-heavy class mixtures, recording the class proportions actually obtained. These experiments change sample composition while keeping documents intact. They do not create new language patterns, topics, annotation rules, or time periods.

\subsection{RQ4 Select Support: Development-to-Test Policy Generalisation}
\label{sec:policy-transfer-results}

\begin{figure*}[!t]
\centering
\includegraphics[width=.98\textwidth]{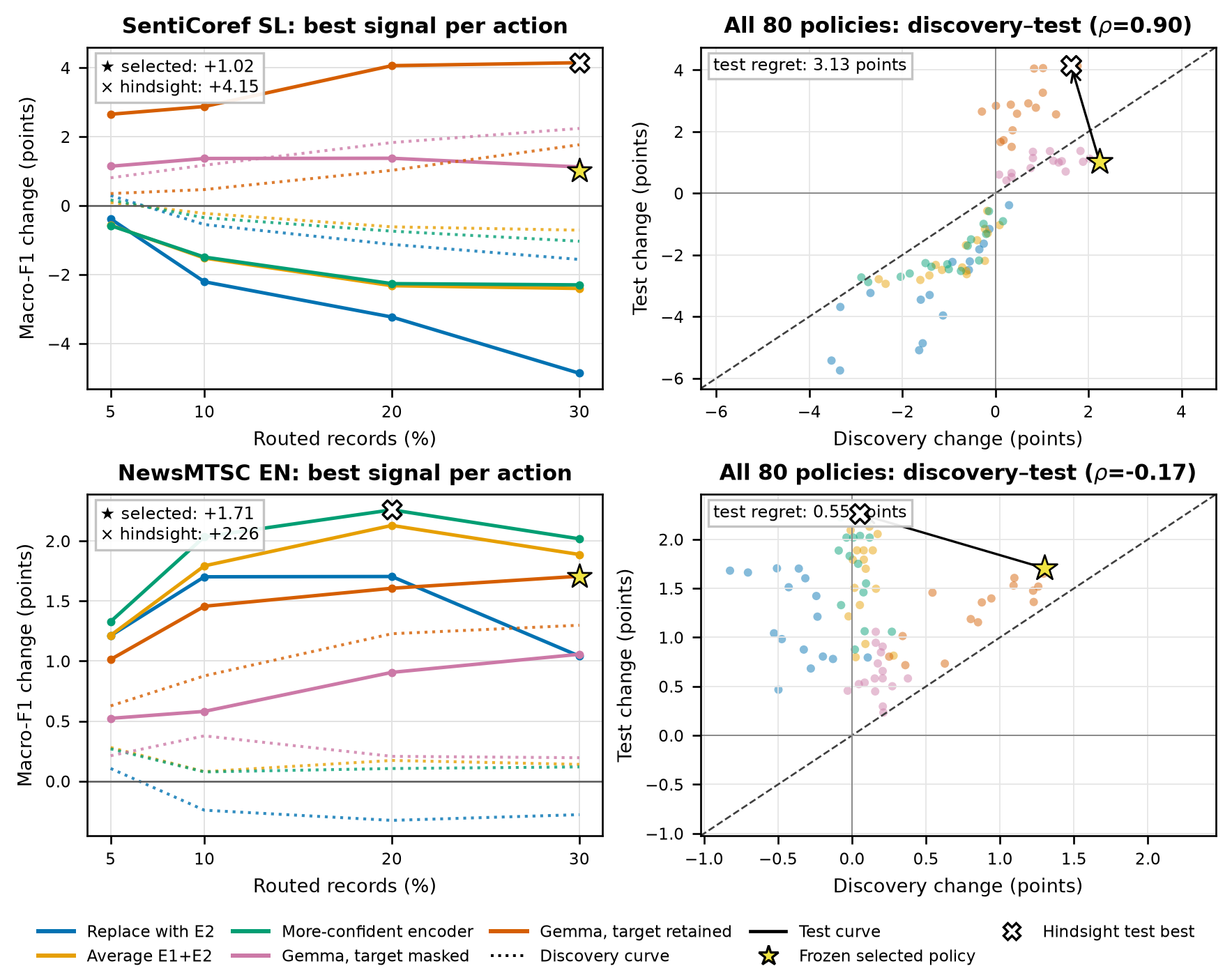}
\caption{Development-selected policies on held-out test data. Dotted curves show the best discovery-set signal for each action/rate and solid curves the retrospective test envelope; stars are frozen choices and crosses hindsight test-best grid members. SentiCoref selects masked Gemma at 30\% ($+1.02$ test points) and NewsMTSC target-retaining Gemma at 30\% ($+1.71$). Hindsight gains of $+4.15$ and $+2.26$ leave regrets of $3.13$ and $0.55$. Strong discovery--test rank correlation on SentiCoref does not ensure the test optimum, while NewsMTSC's weak negative correlation still yields a near-best selected policy.}
\label{fig:policy-transfer}
\end{figure*}

Figure~\ref{fig:policy-transfer} compares all 80 candidates on discovery and test. The left panels show the best discovery-set risk signal for each action and routing rate; the right panels compare discovery and test Macro-F1 changes for every policy. SentiCoref's rankings agree closely across discovery and test ($\rho=.903$), although its selected policy is not the test optimum. NewsMTSC has a weak negative correlation ($\rho=-.169$), yet its selected policy happens to lie close to the best test result. Agreement among whole-policy rankings is therefore not required for one selected point to perform well, nor does strong agreement guarantee selection of the test optimum.

Confirmation checks only the discovery winner. SentiCoref's selected policy gains 2.24 points on discovery, 0.67 on confirmation, and 1.02 on test; the corresponding NewsMTSC gains are 1.30, 0.71, and 1.71. Across all 80 candidates, confirmation--test rank correlations are $-.810$ for SentiCoref and $.754$ for NewsMTSC. Discovery and confirmation also disagree on the candidate rankings ($\rho=-.723$ and $-.494$, respectively). The preferred candidate therefore depends considerably on which development records are used. Confirmation never substitutes its own winner: it only rejects the discovery choice if that policy's observed gain over E1 is non-positive. Both choices pass, although neither matches the hindsight test optimum. Two independent datasets are not enough to treat these outcomes as population estimates.

\subsection{RQ4 Select Support: Contextual Action Routing}
\label{sec:contextual-routing}

\begin{figure*}[!t]
\centering
\includegraphics[width=.95\textwidth]{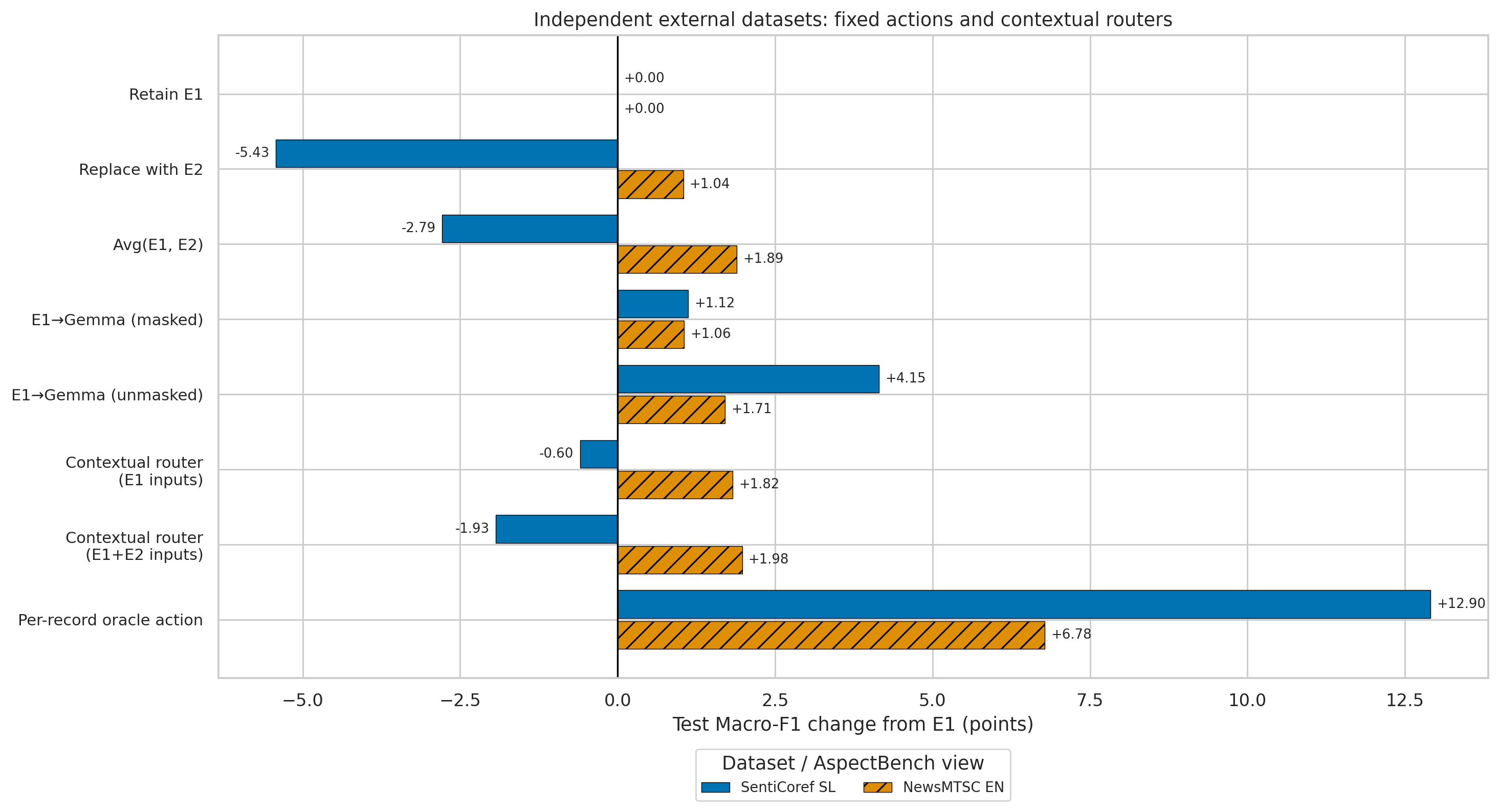}
\caption{Fixed actions and learned per-record routers inside the same lowest-confidence 30\% test region. Blue bars are SentiCoref and hatched orange bars are NewsMTSC. The oracle uses reference labels only to show available headroom. E1 and E1+E2 name the router's input features; E2 replacement, averaging, and Gemma remain candidate actions.}
\label{fig:contextual-actions}
\end{figure*}

On SentiCoref, both learned routers are worse than retaining E1, even though unmasked Gemma helps when applied to every selected record and a label-aware upper bound shows substantial unused opportunity (Figure~\ref{fig:contextual-actions}). On NewsMTSC the router helps, but its +1.98-point gain is only 0.09 points above fixed averaging and requires E2 features for every candidate record.

The cross-dataset diagnostic transfers the five fitted action-correctness models, not primary-model weights or a complete dataset-level policy. For each feature regime, the separate logistic models are fitted on all available source-dataset development candidates and choose among retaining E1, replacing it with E2, averaging, masked Gemma, and target-retaining Gemma within the target dataset's own least-confident 30\%. The E1-input regime uses E1 probabilities, confidence, margin, entropy, mutual information, variation ratio, and predicted-class indicators. The E1+E2 regime additionally requires target-side E2 probabilities and disagreement, even though E2 is a different model on the two datasets.

A SentiCoref-trained E1-input router loses 0.60 Macro-F1 points on SentiCoref, while a NewsMTSC-trained E1-input router gains 1.93 points when applied to SentiCoref; adding the E2 inputs reduces that transferred gain to 0.16. These are single frozen-test point estimates without resampling intervals for router transfer. They show an unexpected positive transfer in one configuration, but do not establish that cross-dataset training is generally superior or that additional features and local fitting guarantee improvement.

AspectBench provides a related but less complete comparison. Out-of-fold development predictions exist for HBS, SL, SR, SH, HR, SR-CYR, and SR-LTN, but these language and script subsets overlap and are not independent datasets. We exclude the Bosnian test slice because it contains only 15 records, has no negative example, and lacks comparable predictions from a separately trained E2. The E1-input router helps three of the seven reported subsets and harms four; the E1+E2 version helps four and harms three. Simple fixed actions often match or exceed it.

\subsection{RQ4 Select Support: Frozen-policy Robustness}
\label{sec:policy-robustness}

\begin{figure*}[!t]
\centering
\includegraphics[width=.95\textwidth]{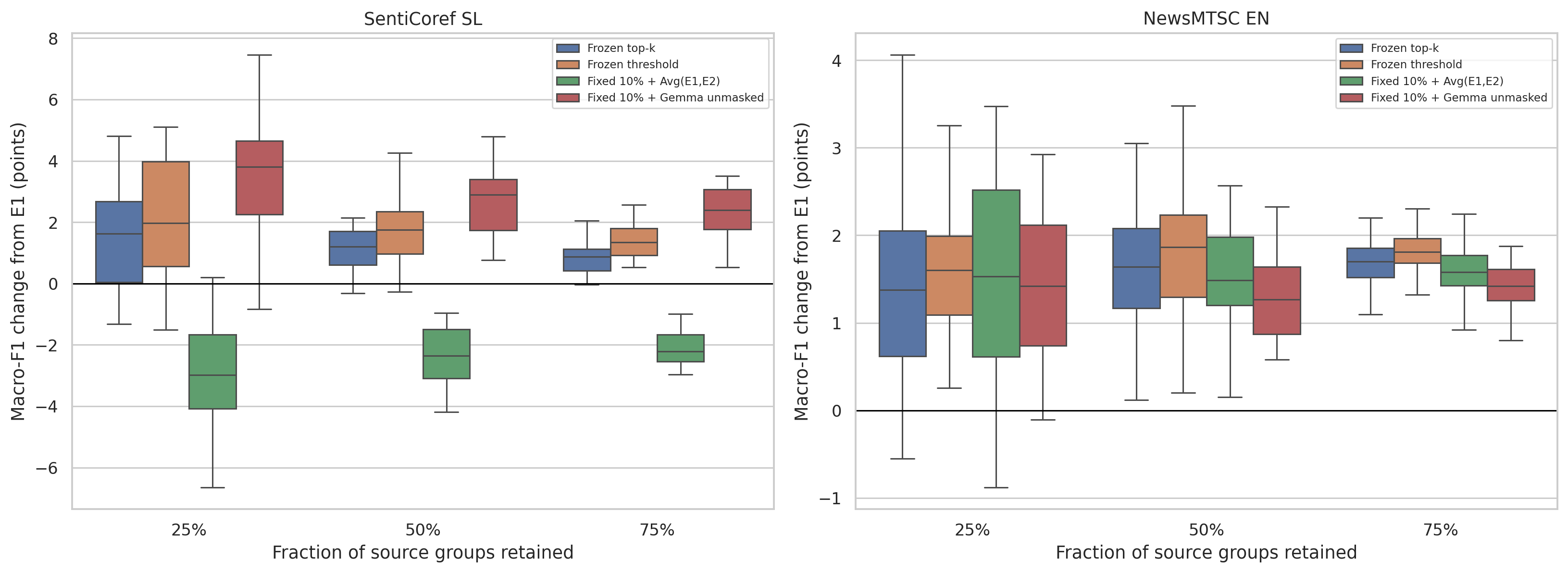}\par\vspace{.4ex}
\includegraphics[width=.95\textwidth]{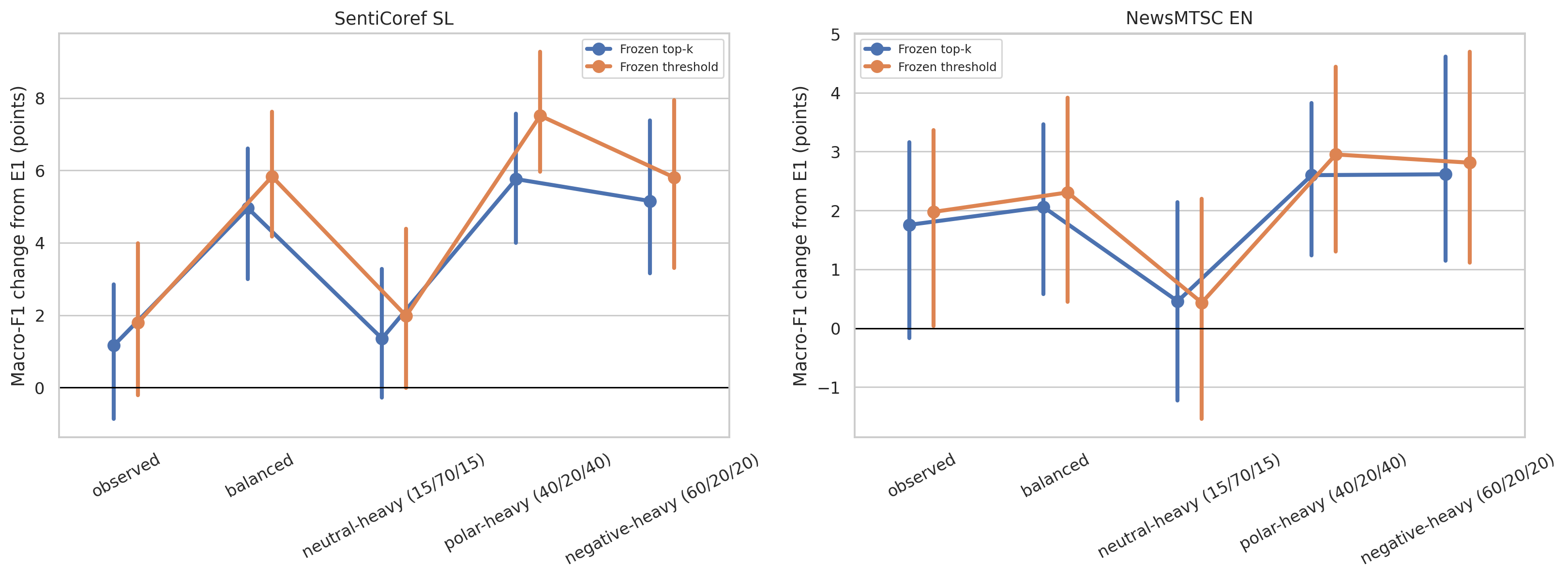}
\caption{Two sensitivity analyses for the frozen development-selected policy. Top: Macro-F1 changes over 30 predetermined source-group-aware subsets at three sample sizes; small SentiCoref samples can reverse the gain, and variability contracts as more groups are retained. Bottom: five controlled class-mixture scenarios with 30 source-group bootstrap replicates each. The numerical threshold often routes more than the fixed 30\% top-$k$ policy, so its gain is not a same-budget comparison. Both panels resample existing test records rather than representing independent deployments or new language, topic, or concept drift.}
\label{fig:heldout-robustness}
\end{figure*}

Keeping the fixed-fraction policy unchanged, SentiCoref's mean gains at 25\%, 50\%, and 75\% retained source groups are +1.57, +1.14, and +0.85 points. The policy hurts performance in 26.7\%, 13.3\%, and 3.3\% of those resamples. NewsMTSC averages +1.48, +1.64, and +1.66 and is harmful in 6.7\%, 0\%, and 0\%. Figure~\ref{fig:heldout-robustness} therefore measures sensitivity to sample size around the existing test result, not performance across new deployments.

Under the observed, balanced, neutral-heavy, polar-heavy, and negative-heavy scenarios, SentiCoref's frozen fixed-fraction policy averages +1.16, +4.95, +1.35, +5.76, and +5.15 points; NewsMTSC averages +1.76, +2.06, +0.46, +2.60, and +2.61. Among the four controlled-prior scenarios, neutral-heavy has the lowest mean on both datasets and is harmful in 10.0\% of SentiCoref and 26.7\% of NewsMTSC resamples. It is not universally the least stable condition: SentiCoref's observed-mixture samples have the lower mean and are harmful in 13.3\%. Moreover, the scenario's approximately 70\% neutral target is below SentiCoref's observed neutral share, so its name identifies the common scenario rather than a dataset-specific shift direction. On the complete tests, the frozen numerical threshold routes 42.83\% of SentiCoref and 38.13\% of NewsMTSC rather than the fixed 30\%. Across the controlled SentiCoref mixtures it routes roughly 46\% of neutral-heavy, 69\% of balanced, 72\% of negative-heavy, and 78\% of polar-heavy records. Throughput planning must therefore track the actual fraction routed, not only the configuration chosen on development data.

NewsMTSC retains two sufficiently large source groups. The frozen fixed-fraction policy gains +0.65 points for AllSides and +2.52 for PolUSA, showing that its value varies by source. SentiCoref has no comparable source field, and neither external dataset has reliable timestamps for at least 80\% of test records. We therefore report no chronological result and do not treat file order or numeric IDs as time.

\end{document}